\documentclass[journal]{IEEEtran}

\usepackage{cite}
\usepackage{amsmath,amssymb}
\usepackage{graphicx}
\usepackage{moreverb,url}
\usepackage[colorlinks,bookmarksopen,bookmarksnumbered,citecolor=red,urlcolor=red]{hyperref}
\usepackage[linesnumbered, ruled]{algorithm2e}
\usepackage{subfig}
\usepackage{multirow}

\newcommand{\etc}{\textit{etc}}
\newcommand{\ie}{\textit{i}.\textit{e}., }
\newcommand{\eg}{\textit{e}.\textit{g}., }
\newcommand{\IR}{\mathbb{R}}

\newcommand{\SE}{\text{SE}}
\newcommand{\SO}{\text{SO}}

\begin{document}

\title{Efficient Path Planning in Narrow Passages for Robots with Ellipsoidal Components}

\author{Sipu Ruan, Karen L. Poblete, Hongtao Wu, Qianli Ma and Gregory S. Chirikjian*
\thanks{Parts of materials from this article were presented in 2018 International Workshop on Algorithmic Foundations of Robotics (WAFR), Merida, Mexico \cite{ruan2018path}.}
\thanks{Sipu Ruan (\texttt{ruansp@nus.edu.sg}) and Gregory S. Chirikjian (\texttt{mpegre@nus.edu.sg}) are with Department of Mechanical Engineering, National University of Singapore, Singapore.}
\thanks{Hongtao Wu is with Laboratory for Computational Sensing and Robotics, Johns Hopkins University, Baltimore, MD 21218, USA.}
\thanks{Karen L. Poblete is with Epic Systems, Verona, WI 53593, USA.}
\thanks{Qianli Ma is with Motional Inc., Pittsburgh, PA 15207, USA.}
\thanks{* Address all correspondence to this author.}}

\maketitle

% Abstract
\begin{abstract}
    Path planning has long been one of the major research areas in robotics, with PRM and RRT being two of the most effective classes of planners. Though generally very efficient, these sampling-based planners can become computationally expensive in the important case of ``narrow passages''. This paper develops a path planning paradigm specifically formulated for narrow passage problems. The core is based on planning for rigid-body robots encapsulated by unions of ellipsoids. Each environmental feature is represented geometrically using a strictly convex body with a $\mathcal{C}^1$ boundary (\eg superquadric). The main benefit of doing this is that configuration-space obstacles can be parameterized explicitly in closed form, thereby allowing prior knowledge to be used to avoid sampling infeasible configurations. Then, by characterizing a tight volume bound for multiple ellipsoids, robot transitions involving rotations are guaranteed to be collision-free without needing to perform traditional collision detection. Furthermore,  by combining with a stochastic sampling strategy, the proposed planning framework can be extended to solving higher dimensional problems in which the robot has a moving base and articulated appendages. Benchmark results show that the proposed framework often outperforms the sampling-based planners in terms of computational time and success rate in finding a path through narrow corridors for both single-body robots and those with higher dimensional configuration spaces. Physical experiments using the proposed framework are further demonstrated on a humanoid robot that walks in several cluttered environments with narrow passages.
\end{abstract}


%%%%%%%%%%%%%%%%%%%%%%%%%%%%%%%%%%%%%%%%%%%%%%%%%%%%%%%%%%
% Introduction
\section{Introduction}
\label{sec:introduction}
Sampling-based planners such as PRM \cite{kavraki1996probabilistic} and RRT \cite{lavalle1998rapidly} (and a multitude of their extensions, \eg \cite{kuffner2000rrt,bohlin2000path}) have demonstrated remarkable success in solving complex robot motion planning problems. These frameworks generate state samples randomly and perform explicit collision detection to assess their feasibility. These methods have had a profound impact both within robotics and across other fields such as molecular docking, urban planning, and assembly automation.

It is well known that despite the great success of these methods, the ``narrow passage'' problem remains a significant challenge. Generally speaking, when there is a narrow passage, an inordinate amount of computational time is spent on the random state samples and edges that eventually will be discarded. To increase the probability of sampling and connecting valid configurations in a narrow passage, various methods have been proposed such as \cite{hsu2003bridge,shi2014spark,lai2020bayesian} (Sec. \ref{sec:related-work:narrow_passage} provides more detailed reviews on narrow passage problems). In this article, however, the narrow passage problem is addressed through an explicit closed-form characterization of the boundary between free and in-collision regions. The first goal of this paper is to:

{\it 1. Extend the previous methods of parameterizing the free space for single-body ellipsoidal robots avoiding ellipsoidal obstacles \cite{yan2016path}. A more general case is studied where the obstacles are represented by unions of strictly convex bodies with $\mathcal{C}^1$ boundaries.}

\begin{figure}[!t]
\centering
\subfloat[Projection contour of a NAO humanoid robot is enclosed by an ellipse (yellow).]{
\label{fig:demo_ellipsoid_robot:nao_ellipse}
\includegraphics[scale=0.25]{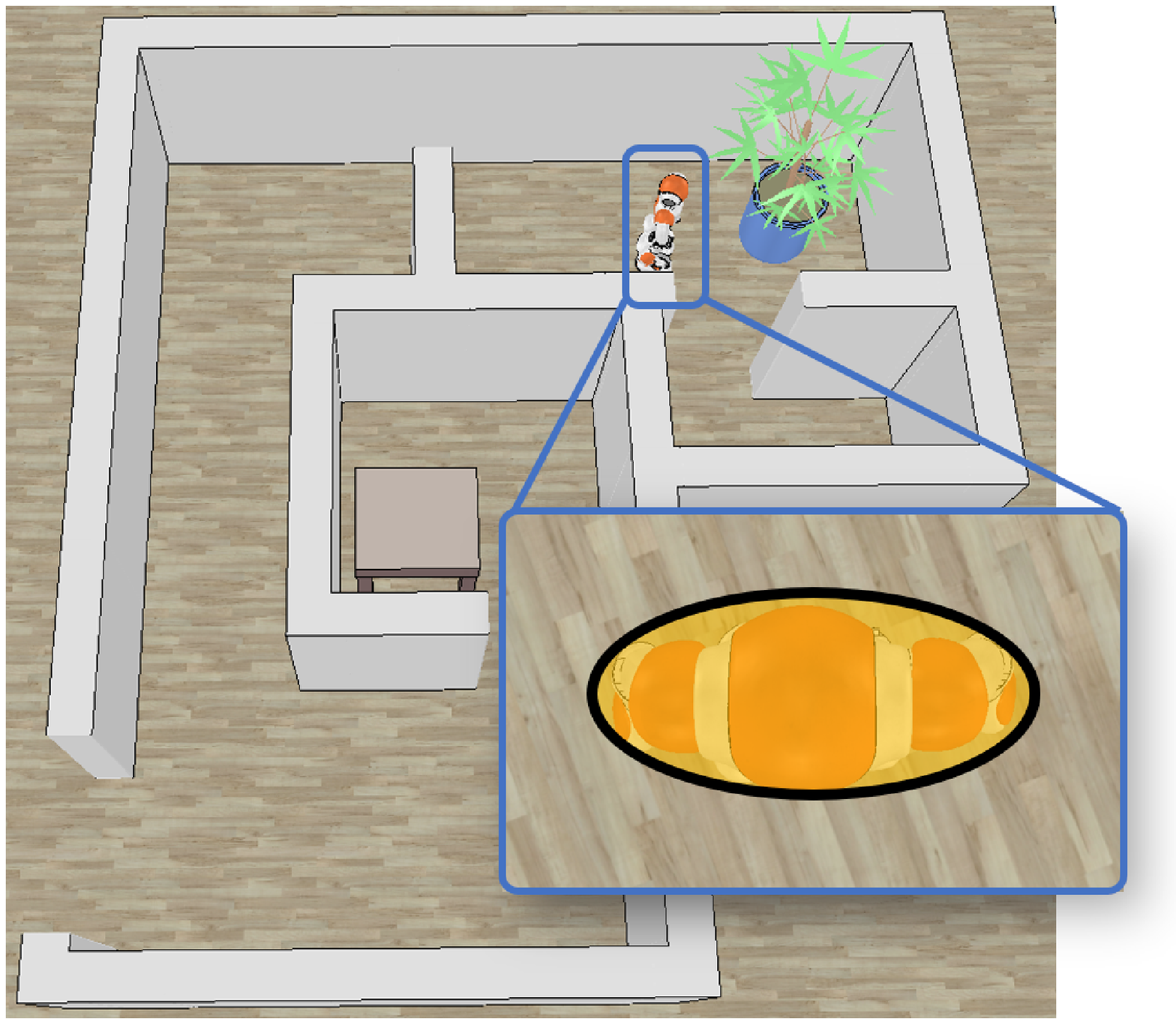}}
\hspace{0.1in}
\subfloat[Protein atom elements (small balls) are enclosed by ellipsoids (green).]{
\label{fig:demo_ellipsoid_robot:protein_ellipsoid}
\includegraphics[scale=0.24]{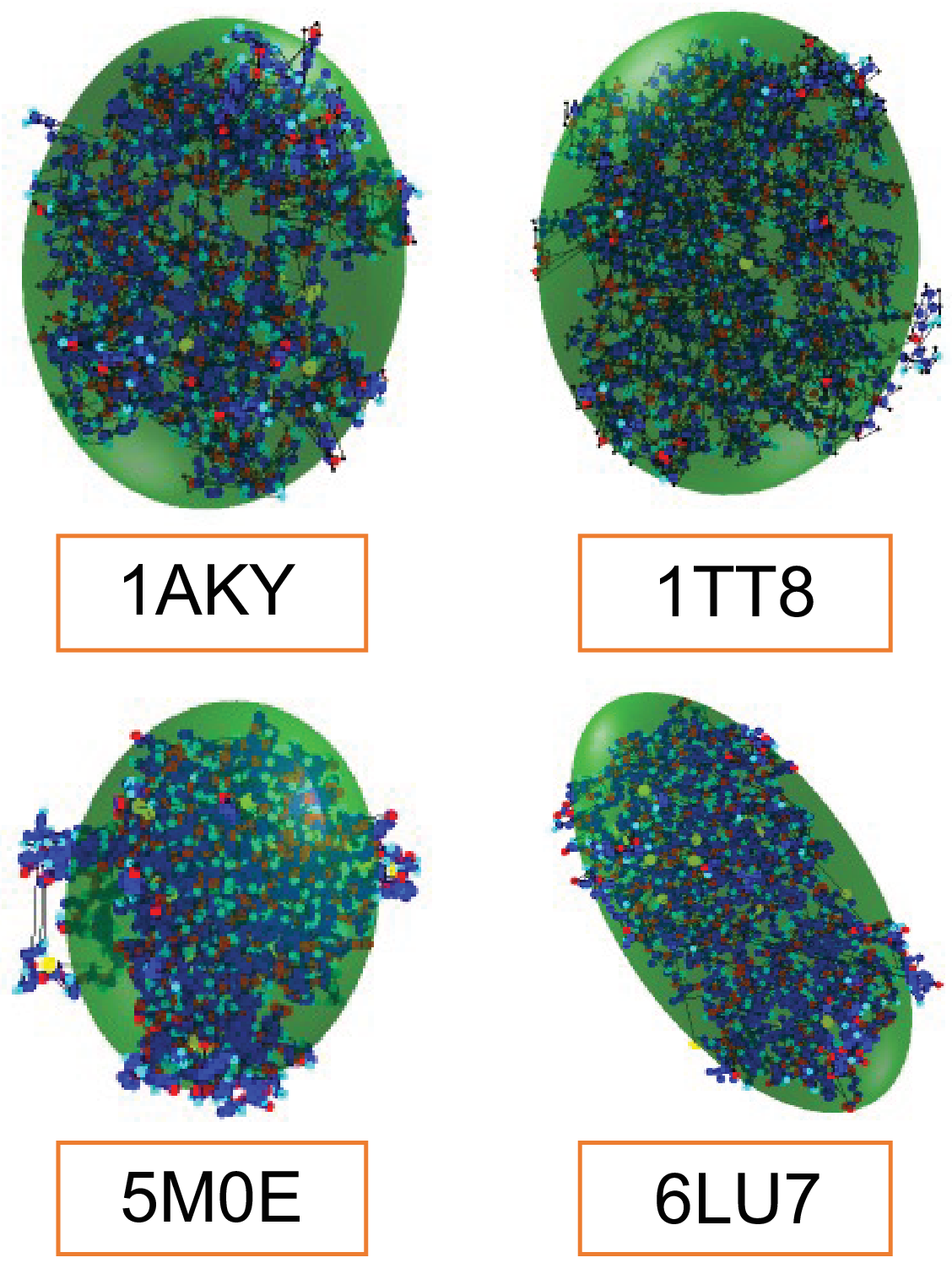}}
\caption{Examples of robots and protein molecules encapsulated by ellipsoids.}
\label{fig:demo_ellipsoid_robot}
\end{figure}

In our proposed path planning framework, the robot is encapsulated by a union of ellipsoids. The configuration spaces to be considered are $\SE(d)$ and $\SE(d) \times (S^1)^n$ for rigid-body and articulated robots, respectively\footnote{$\SE(d), \, d = 2,3$ is the pose of the robot base frame and $(S^1)^n$ represents the configuration space of $n$ revolute joints.}. Ellipsoids have a wide range of applications in encapsulating robots. For example, the projection contour of a humanoid robot can be tightly encapsulated by an ellipse since its shoulders are narrower than the head \cite{best2016real} (Fig. \ref{fig:demo_ellipsoid_robot:nao_ellipse}). In computational crystallography, it is natural to approximate a protein molecule by a moment-of-inertia ellipsoid, which simplifies the complex geometric models and maintains the physical information of the protein \cite{shiffman2020mathematical} (Fig. \ref{fig:demo_ellipsoid_robot:protein_ellipsoid}). Moreover, superquadrics are chosen as examples to represent environmental features. This family of shapes generalizes ellipsoids by adding freedoms in choosing the power of the exponents rather than restricting to quadratics. It represents a wider range of the complex shapes (\eg cuboids, cylinders, \etc) while also requiring only a few parameters \cite{barr1981superquadrics}. 

When a robot is fixed at a certain orientation and internal joint angles, a ``slice'' of the configuration space (C-space) is defined by the Minkowski sums between the rigid body parts and the obstacles in the workspace \cite{latombe2012robot,nelaturi2011configuration}, denoted here as a ``C-slice'' \cite{lien2008hybrid}. (Sec. \ref{sec:related-work:minkowski} reviews the literature in details on the computations of Minkowski sums). Once the C-space obstacles (C-obstacles) are computed, the complement regions between the planning arena\footnote{Here, the word ``arena'' denotes the bounded area in which the robot and obstacles are contained.} and the union of C-obstacles is the free space that allows the robot to travel through safely. Consequently, collision-free samples can be generated within this collision-free C-space. However, if one seeks to {\it connect} such samples using current sampling-based planners like PRM or RRT, explicit collision checking is still required. Therefore, the second goal of this article is to:

{\it 2. Develop guaranteed safe and efficient methods for connecting configurations between different C-slices \underline{without} performing explicit collision checking between pairwise bodies.}

A ``bridge C-slice'' idea is proposed as a local planner to guarantee safe transitions between different C-slices. The name suggests that a new C-slice is built as a bridge between two adjacent C-slices. To efficiently construct a bridge C-slice, an enlarged void for each ellipsoidal robot part is computed in closed form. Here, a ``void'' is the free space that fully contains the robot part, ensuring that it moves without collisions. A sweep volume is then constructed to enclose the robot at all possible intermediate configurations during the transition. 

All the above methods are combined into a path planning algorithm called ``Highway RoadMap (HRM)''. This planner is deterministic and suitable for rigid-body planning problems. It is known that traditional deterministic planners suffer from the curse of dimensionality burden in the case of articulated robots. Therefore, the third goal of this article is to:

{\it 3. Develop an effective method to tackle the exponential computational complexity for the planning of articulated robots.}

A hybrid algorithm called ``Probabilistic Highway RoadMap (Prob-HRM)'' is proposed here to make planning in higher dimensional configuration spaces tractable. It randomly samples the rotational components (\ie the base orientation and internal joint angles) and takes advantage of the explicit parameterizations of free space in each C-slice from HRM.

This article extends the conference version \cite{ruan2018path} on the same topic, and has significant updates. Comparing to the conference paper, the key contributions of this article are:
\begin{itemize}
\item Extend the graph construction procedure in each C-slice to 3D multi-body case;
\item Introduce a novel ``bridge C-slice'' method to connect vertices between adjacent C-slices;
\item Propose a hybrid planner which integrates the advantages of sampling-based planners on higher dimensional articulated robot planning problems;
\item Conduct rigorous benchmark simulations and physical experiments in challenging environments to evaluate the proposed planning framework.
\end{itemize}
These extensions are essential since more general 3D and articulated robot models are implemented. The benchmark and physical experimental settings are also more realistic.

The rest of this article is organized as follows. Section \ref{sec:related-work} reviews related literature. Section \ref{sec:math-pre} provides mathematical foundations. Section \ref{sec:hrm-planner} extends our previously proposed HRM planner to the case of 3D multi-body robot with ellipsoidal components. The novel ``bridge C-slice'' method is then introduced. Section \ref{sec:prob-hrm-planner} introduces the hybrid Prob-HRM planner. Section \ref{sec:benchmark} conducts extensive benchmarks with some popular and successful sampling-based planners. In Section \ref{sec:real-experiment}, our planning framework is demonstrated by physical experiments in real world, which solve walking path planning problems for a humanoid robot in cluttered environments. We discuss the advantages and limitations of our proposed framework in Section \ref{sec:discussion}. Finally, we conclude in Section \ref{sec:conclude}.

% Related work
\section{Literature Review}
\label{sec:related-work}
This section reviews related work on the key topics that this article addresses.

%%%%%%%%%%%%%%%%%%%%%%%%%%%%%%%%%%%
\subsection{The challenge of narrow passages}
\label{sec:related-work:narrow_passage}
One of the key factors that affects the performance of sampling-based planners is the random state sampling strategy. To tackle the ``narrow passage'' challenge, various sampling strategies have been studied throughout these years, many of which try to capture the local features around obstacles.

The bridge test \cite{hsu2003bridge} finds a collision-free middle point between configurations that are in collision with the obstacles. UOBPRM \cite{yeh2012uobprm} searches for collision-free samples from a configuration in collision by moving in different ray directions. In \cite{lai2020bayesian}, a Bayesian learning scheme is used to model sampling distributions. It subsequently updates the previous samples by maximizing the likelihood from the region that has higher probability in forming a valid path within the narrow passage. Ideas about generating samples on the ``medial axis'' were proposed in \cite{wilmarth1999maprm, yeh2014umaprm}. Each sampled state, regardless of free or in-collision, is retracted to the medial axis of free space. The retraction direction is selected between the sampled state and its nearest neighbor on the boundary of free space. The resulting samples stay far from obstacles. And the usage of in-collision samples is able to detect regions close to narrow passages. The proposed framework in this article also attempts to generate vertices that stay away from obstacles as far as possible. A similar idea is used in the ``maximize clearance'' sampler, \ie PRM(MC), in the benchmark studies of this article. For each valid sample, the sampler searches a new sample close-by but with larger distance to the obstacles. We use PRM(MC) for comparisons since it is implemented on the well-known Open Motion Planning Library (OMPL) \cite{sucan2012open}. This provides a standardized way to benchmark with other sampling-based planning algorithms as well as samplers.

Other methods combine the advantages of different kinds of algorithms. For example, Toggle PRM \cite{denny2013toggle} simultaneously maps both free space and obstacle space, enabling an augmentation from a failed connection attempt in one space to the other. Spark PRM \cite{shi2014spark} grows a tree inside the narrow passage region to connect different parts of the roadmap on different ends of the region. Retraction-based RRT \cite{lee2014selective} tries to retract initial samples into more difficult regions, so as to increase probability of sampling near narrow passages. More recently, a reinforcement learning method is applied to enhance the ability to explore local regions where the tree grows \cite{wang2018learning}. 

Hybrid planner \cite{lien2008hybrid} combines a random sampling strategy with Minkowski sums computations, which increases the probability of identifying narrow regions. In this article, we use an approach with some similarities to the Prob-HRM planner to randomly sample the robot shapes. Nevertheless, the differences are significant. We propose a closed-form Minkowski sum expression for continuous bodies, as compared to point-based Minkowski sums for polyhedral objects. To generate valid vertices, they directly choose the points on C-obstacle boundary, but we generate in the middle of free space in a more uniform way. And to connect different C-slices, they add a new vertex and search for paths on the C-obstacle boundaries, but we instead generate a new slice based on an enlarged void.

%%%%%%%%%%%%%%%%%%%%%%%%%%%%%%%%%%%
\subsection{Computations of Minkowski sums}
\label{sec:related-work:minkowski}
The Minkowski sum is ubiquitous in many fields such as computational geometry \cite{qin2019smoothing}, robot motion planning \cite{latombe2012robot}, control theory \cite{halder2020smallest}, \etc. Despite its straightforward definition, which will be given in Sec. \ref{sec:math-pre}, computing an exact boundary of Minkowski sum between two general non-convex polytopes in $\IR^3$ can be as high as $O(N_1^3 N_2^3)$, where $N_1$ and $N_2$ are the complexities (\ie the number of facets) of the two polytopes. Therefore, many efficient methods decompose the general polytopes into convex components \cite{hachenberger2009exact}, since the Minkowski sums between two convex polytopes can achieve $O(N_1 N_2)$ complexity \cite{fogel2009exactcomplexity}. Another type of methods is based on convolutions of two bodies, since Minkowski sum of two solid bodies is the support of the convolution of their indicator functions \cite{chirikjian2016harmonic}. A simple approximated algorithm \cite{lien2009simple} is proposed that avoids computing 3D arrangement and winding numbers via collision detection. An exact Minkowski sum for polytopes containing holes is proposed using convolution \cite{baram2018exact}. In addition, point-based methods avoid convex decomposition \cite{barki2009contributing}. The major advantages are the ease of generating points than meshes and the possibility of parallelisms \cite{lien2008covering}. An exact closed-form Minkowski sum formula for $d$-dimensional ellipsoids was introduced \cite{yan2015closed}. And in \cite{halder2018parameterized}, a parameterized ellipsoidal outer boundary for the Minkowski sum of two ellipsoids is proposed. This article studies a more general case when one body is an ellipsoid and the other is a strictly convex body with a $\mathcal{C}^1$ boundary (\eg superquadric).

%%%%%%%%%%%%%%%%%%%%%%%%%%%%%%%%%%%
\subsection{Ellipsoids and superquadrics for object representation}
\label{sec:related-work:ellipsoid}
Besides using polyhedra for object representations, other geometric primitives such as ellipsoids and superquadrics also play an important role due to their simple algebraic characterizations. Recently, in many robotic applications, they are good candidates to encapsulate objects \cite{vezzani2017grasping,polverini2020multi}.

A 3D ellipsoid in a general pose only needs 9 parameters: 3 for the shape (\ie semi-axes lengths) and 6 for the pose. Algorithms related to ellipsoids are studied extensively \cite{pope2008algorithms,kurzhanskiy2006ellipsoidal}. The minimum volume enclosing ellipsoid (MVEE), which is characterized as a convex optimization problem \cite{khachiyan1996rounding}, is widely used to encapsulate a point cloud. The studies of algebraic separation conditions for two ellipsoids provides very efficient algorithms to detect collisions in both static and dynamic cases \cite{wang2001algebraic,jia2011algebraic}. Another attractive attribute of the representation using ellipsoids is the existence of efficient procedures of computing their distance \cite{rimon1997obstacle,iwata2015computing}. Once an ellipsoid is fully contained in another, the volume of its limited available motions is computed explicitly \cite{ruan2019kinematics}.

Superquadrics can be seen as an extension of ellipsoids, with the two additional exponents determining the sharpness and convexity \cite{barr1981superquadrics}. They are able to represent a wider range of geometries such as cube, cylinder, octahedron, \etc. Using optimization or deep learning techniques, point cloud data can be segmented and fitted by unions of superquadrics \cite{vaskevicius2017revisiting,paschalidou2019superquadrics}. Proximity queries and contact detection are useful applications of this geometric model \cite{peng2019contact,ruan2019efficient}.

% Math Preliminary
\section{Mathematical Preliminaries}
\label{sec:math-pre}
This section provides the mathematical preliminaries for developing the new path planning paradigm in this article.

%%%%%%
\subsection{Minkowski sum and difference between two bodies} \label{sec:math:mink-general}
The Minkowski sum and difference of two point sets (or bodies) centered at the origin, \ie $P_1$ and $P_2$ in $\IR^d$, are defined respectively as \cite{berg2008computational}
\begin{equation}
\begin{aligned}
P_1 \oplus P_2 & \,\doteq\, \{p_1+p_2~|~p_1 \in P_1, p_2 \in P_2 \}, && \text{and} \\
P_1 \ominus P_2 & \,\doteq\, \{p~|~p + P_2 \subseteq P_1 \}.
\end{aligned}
\end{equation}
When computing the boundary in which the two bodies touch each other externally (\ie their contact space), we refer to the calculation of $\partial [P_1 \oplus (-P_2)]$, where $-P_2$ is the reflection of $P_2$ as viewed in its body frame \cite{lien2009simple}. Note that when $P_2$ is centrally symmetric, such as ellipsoids and superquadrics that this article focuses on, the Minkowski sum boundary and contact space are equivalent. Moreover, when the bodies are non-convex, using the fact that
\begin{equation}
\text{if} \,\, P_1 = Q_1 \cup Q_2, \,\, \text{then,} \,\, P_1 \oplus P_2 = (Q_1 \oplus P_2) \cup (Q_2 \oplus P_2),
\label{eq:math:union_mink}
\end{equation}
their Minkwoski sums can be obtained via convex decomposition.

%%%%%%
\subsection{Implicit and parametetric surfaces}
Assume that $S_1$ is a strictly convex body bounded by a $\mathcal{C}^1$ hyper-surface embedded in $\IR^d$. The implicit and parametric forms of its surface can be expressed as
\begin{equation}
\Phi({\bf x}_1) = 1 \,\, \text{and} \,\, {\bf x}_1 = {\bf f}(\boldsymbol{\psi}_1),
\end{equation}
where $\Phi(\cdot)$ is a real-valued differentiable function of ${\bf x}_1 \in \IR^d$ and ${\bf f}$ is a differentiable $d$-dimensional vector-valued function of surface parameters $\boldsymbol{\psi}_1 = [\psi_1, \psi_2,..., \psi_{d-1}]^\top \in \IR^{d-1}$.

Let $E_2$ be an ellipsoid in $\IR^d$ in general orientation, with semi-axis lengths ${\bf a}_2 = [a_1, a_2,...,a_n]^\top$. Its implicit and explicit expressions are
\begin{equation}
{\bf x}_2^\top A_2^{-2} {\bf x}_2 = 1 \,\, \text{and} \,\, {\bf x}_2 = A_2 {\bf u}(\boldsymbol{\psi}_2),
\label{eq:math:ellipsoid_model}
\end{equation}
where $A_2 = R_2 \Lambda({\bf a}_2) R_2^\top$ is the shape matrix of $E_2$, $R_2 \in \SO(d)$ denotes the orientation of $E_2$, and $\Lambda(\cdot)$ is a diagonal matrix with the semi-axis length $a_i$ at the $(i,i)$ entry. $A_2^{-2} \doteq (A_2^2)^{-1} = (A_2^{-1})^2$ is used here for the sake of simplicity. ${\bf u}(\boldsymbol{\psi}_2)$ is the standard parameterization of the $d$-dimensional unit hyper-sphere using $d-1$ angles. Specifically, in 2D, ${\bf u}(\theta) = [\cos \theta, \sin \theta]^\top$, and in 3D, ${\bf u}(\eta, \omega) = [\cos \eta \, \cos \omega, \, \cos \eta \, \sin \omega, \, \sin \eta]^\top$.

One class of strictly convex bodies meeting the conditions stated earlier include those with specific kinds of superquadric boundaries. The implicit equations in the 2D and 3D cases are given by
\begin{equation}
\Phi(x,y) = \left( \frac{x}{a} \right)^{\frac{2}{\epsilon}} + \left( \frac{y}{b} \right)^{\frac{2}{\epsilon}} \,\, \text{and}
\label{eq:math:sq_2d}
\end{equation}
\begin{equation}
\Phi(x,y,z) = \left( \left( \frac{x}{a} \right)^{\frac{2}{\epsilon_2}} + \left( \frac{y}{b} \right)^{\frac{2}{\epsilon_2}} \right)^{\frac{\epsilon_2}{\epsilon_1}} + \left( \frac{z}{c} \right)^{\frac{2}{\epsilon_1}} \,, 
\label{eq:math:sq_3d}
\end{equation}
where $a,b,c$ are the semi-axes lengths, and $\epsilon$, $\epsilon_1, \epsilon_2 \in (0,2)$ are the exponents that ensure strict convexity.

%%%%%%
\subsection{Closed-form Minkowski operations between an ellipsoid and a general convex differentiable surface} \label{sec:math:closed-form-mink}
\begin{figure*}[!t]
\centering
\subfloat[The original space with $S_1$ in the center and $E_2$ translating around.]{
  \label{fig:subfig:a}
  \centering
  \includegraphics[scale=0.75]{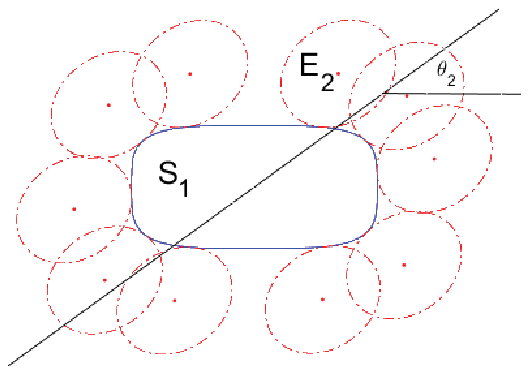}}
\hspace{0.1in}
\subfloat[Both bodies are rotated by the inverse orientation of $E_2$.]{
  \label{fig:subfig:b}
  \centering
  \includegraphics[scale=0.75]{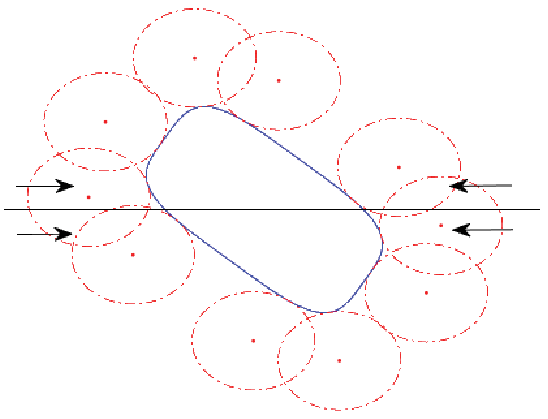}}
\hspace{0.1in}
\subfloat[$E_2$ is shrunk into a sphere, and an offset surface is computed.]{
  \label{fig:subfig:c}
  \centering
  \includegraphics[scale=0.75]{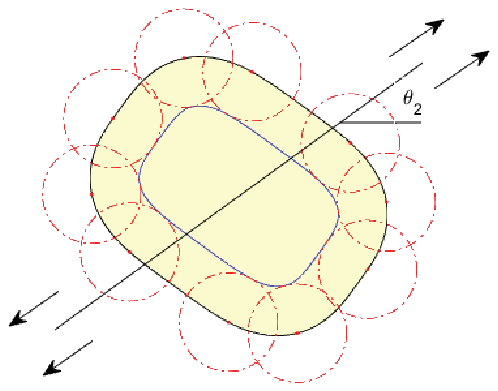}}
\hspace{0.1in}
\subfloat[Stretch back and obtain $S_1 \oplus (-E_2)$ (the yellow region).]{
  \label{fig:subfig:d}
  \includegraphics[scale=0.75]{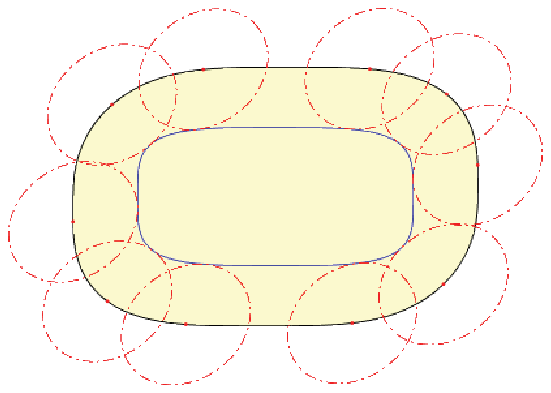}}
\caption{Process for the characterizations of the Minkowski sums between a superquadric $S_1$ and an ellipsoid $E_2$.}
\label{fig:M-sum}
\end{figure*}

It has been shown previously in \cite{yan2015closed} that the Minkowski sum and difference between two ellipsoids can be parameterized in closed-form. The expression can be extended when one ellipsoid is substituted by $S_1$ \cite{ruan2018path}. The general simplified form for the Minkowski sum can be computed as
\begin{equation}
{\bf x}_{mb} = {\bf x}_1 + \frac{R_2 \Lambda^2({\bf a}_2) R_2^\top \nabla_{{\bf x}_1} \Phi({\bf x}_1)}{\| \Lambda({\bf a}_2) R_2^\top \nabla_{{\bf x}_1} \Phi({\bf x}_1) \|},
\label{eq:math:mink_bound_general_simplified}
\end{equation}
where $\nabla_{{\bf x}_1} \Phi({\bf x}_1)$ is the gradient of $S_1$ at ${\bf x}_1$. The conditions that $S_1$ is strictly convex and its boundary is $\mathcal{C}^1$ ensure that the gradient exists and that there is never division by zero when using Eq. \eqref{eq:math:mink_bound_general_simplified}. Figure \ref{fig:M-sum} illustrates the geometric interpretation of the computational process. Detailed derivations were presented in \cite{ruan2018path}.

%%%%%%
\subsection{The minimum volume concentric ellipsoid (MVCE) enclosing two ellipsoids with the same center} \label{sec:math:mvce}
When two ellipsoids are fixed at the same center, a ``minimum volume concentric ellipsoid (MVCE)'' can be computed in closed form as follows. 

\begin{figure}[!t]
\centering
\subfloat[Two concentric 3D ellipsoids, $E_a$ and $E_b$.]{\includegraphics[scale = 0.26, trim = 100 0 80 0, clip]{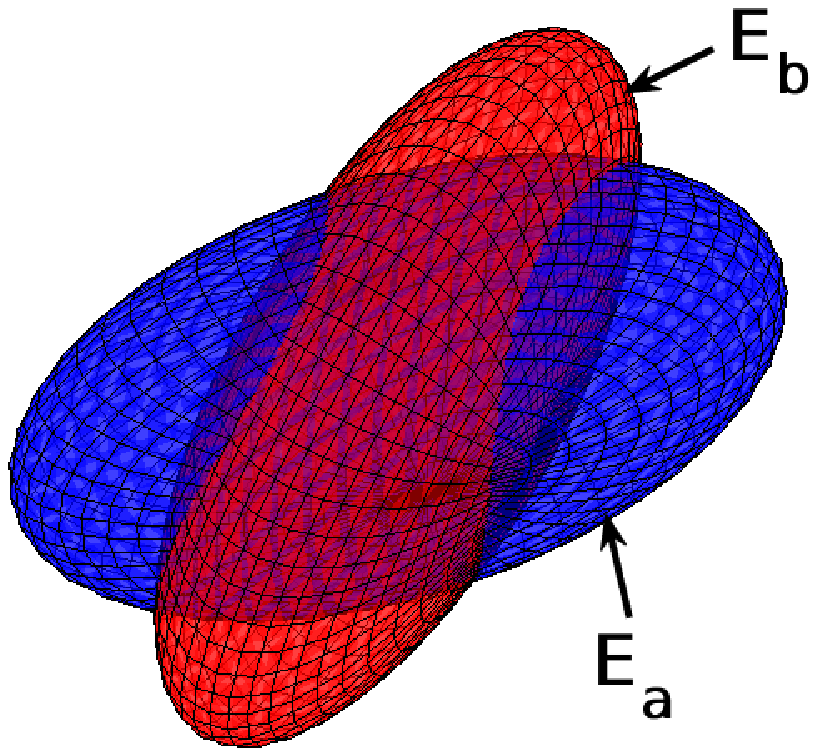}}
\hspace{0.1in}
\subfloat[Shrink $E_b$ into a sphere $E'_b$, and enclose both ellipsoids by $E'_m$.]{\includegraphics[scale = 0.26, trim = 60 20 60 10, clip]{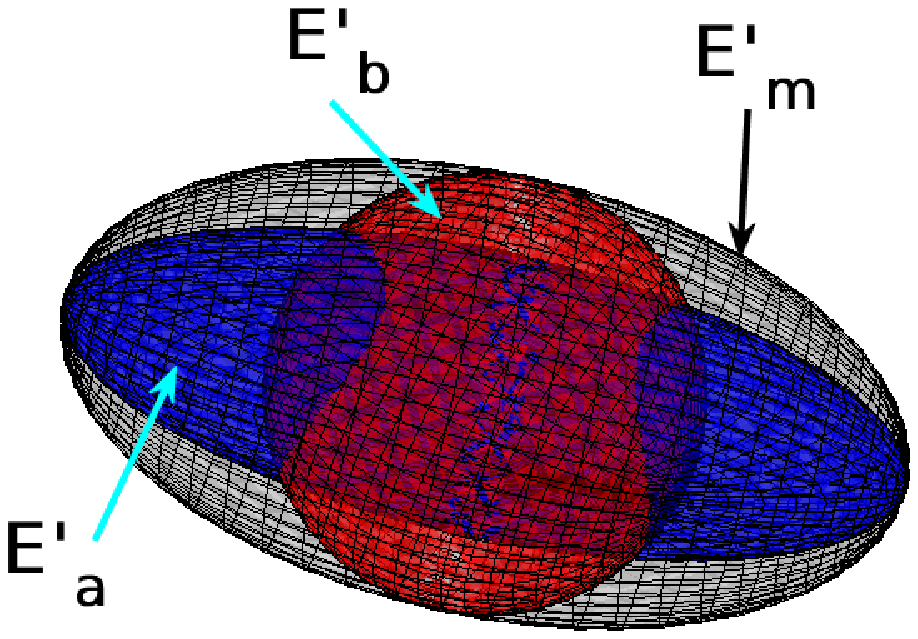}}
\hspace{0.1in}
\subfloat[Transform back to get MVCE $E_m$ in original space.]{\includegraphics[scale = 0.26, trim = 60 20 70 20, clip]{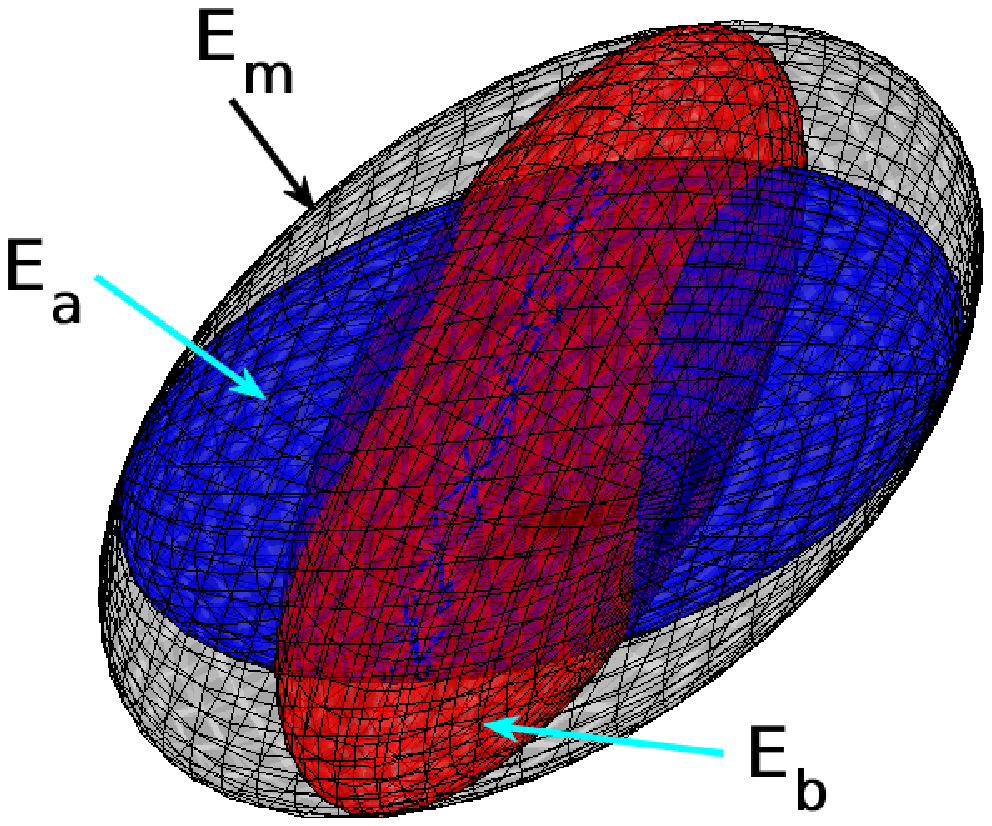}}
\caption{Computational procedure for minimum volume concentric ellipsoid that covers two ellipsoids in 3D.}
\label{fig:mvce_3d}
\end{figure}

Given two $d$-dimensional ellipsoids $E_a$ and $E_b$ with semi-axis lengths ${\bf a}$ and ${\bf b}$ respectively. One ellipsoid (\eg $E_b$) can be shrunk into a sphere ($E'_b$) via the affine transformation $ T = R_b \Lambda(r/{\bf b}) R^\top_b $, where $r$ is the radius and $r/{\bf b} \doteq [r/b_1, \, r/b_2, \, ..., \, r/b_d]^\top \in \IR^d$. Then the shape matrix of $E_a$ in shrunk space, \ie $E'_a$, can be computed as $ A' = T^{-1} R_a \Lambda^{-2}({\bf a}) R^\top_a T^{-1} $. Using singular value decomposition (SVD), its semi-axis lengths and orientation, \ie ${\bf a}'$ and $R'_a$, can be obtained respectively. The shape matrix of their MVCE, \ie $E_m$, is obtained as $ M = T R'_a \Lambda^{-2}(\max({\bf a}', r)) R'^\top_a T $, where $\max({\bf a}', r) \doteq [\max(a'_1,r), \, ..., \, \max(a'_d,r)]^\top$ and ${\bf a}' \doteq [a'_1, \, a'_2, \, ..., \, a'_d]^\top \in \IR^d$. The computational procedure is visualized in Fig. \ref{fig:mvce_3d} for the 3D case. The idea here is inspired by \cite{pope2008algorithms}, which provides equivalent computations for a maximum volume concentric ellipsoid covered by two ellipsoids.

% , their shape matrices can be expressed as $A = R_a \Lambda^{-2}({\bf a}) R^\top_a$ and $B = R_b \Lambda^{-2}({\bf b}) R^\top_b$

Furthermore, this process can be applied iteratively if there are multiple concentric ellipsoids. For example, the MVCE that enclose the previous two ellipsoids, along with the next ellipsoid, can be enclosed by a new MVCE. The final resulting ellipsoid encapsulates all the original set of ellipsoids, which is denoted as a tightly-fitted ellipsoid (TFE).

%%%%%%
\subsection{Superquadric model fitting to point cloud data}
Given a set of $m$ 3D points \{${\bf x}_i = [x_i,y_i,z_i]^\top, \,\, i=1,...,m\}$, a superquadric model can be approximated by minimizing \cite{solina1990recovery,paschalidou2019superquadrics}
\begin{equation}
\min_{a,b,c,\epsilon_1,\epsilon_2,R,{\bf t}} abc \sum_{i=1}^m \left( \Phi^{\epsilon_1}(x'_i,y'_i,z'_i) - 1 \right)^2 \,,
\label{eq:sq_fitting_3d}
\end{equation}
where $\Phi(\cdot)$ is shown in Eq. \eqref{eq:math:sq_3d}, ${\bf x}'_i = R^\top ( {\bf x}_i - {\bf t} )$ is the transformed data point as viewed in the body frame of the superquadric, and $R \in \SO(3)$ and ${\bf t} \in \mathbb{R}^3$ are the orientation and center of the superquadric, respectively. The factor $abc$ is added here in order to minimize the volume of the fitted superquadric body. Similarly, for the 2D case, the corresponding nonlinear optimization problem can be formulated as
\begin{equation}
\min_{a,b,\epsilon,\theta,{\bf t}} ab \sum_{i=1}^m \left( \Phi^{\epsilon}(x'_i,y'_i) - 1 \right)^2 \,,
\label{eq:sq_fitting_2d}
\end{equation}
where $\Phi(\cdot)$ is now referred to Eq. \eqref{eq:math:sq_2d} and $\theta$ is the rotational angle of the 2D superellipse.

Solving the above optimizations requires good initial conditions. The parameters from minimum volume enclosing ellipsoid (MVEE) are used, which can be computed using convex optimization as \cite{khachiyan1996rounding}
\begin{equation}
\begin{aligned}
\min_{A, {\bf t}} & \log \det A \\
{\rm s.t. \,\,} & A \succ 0 \,, \\
& ({\bf x}_i - {\bf t})^\top A^{-2} ({\bf x}_i - {\bf t}) \leq 1 \,\, (i = 1,...,m) \,,
\end{aligned}
\end{equation}
where $A$ is the shape matrix of an ellipsoid as in Eq. \eqref{eq:math:ellipsoid_model}. This convex optimization process can also be used to bound the robot parts if they are originally modeled by surface meshes.

% Highway Roadmap planner
\section{Highway RoadMap Planning Algorithm for Rigid-body Robots with Ellipsoidal Components}
\label{sec:hrm-planner}
This section introduces the extended ``Highway RoadMap (HRM)'' algorithm. The extension to the previous work \cite{yan2016path} from 2D to 3D rigid-body path planning problems is explained here. Then, a novel vertex connection strategy for configurations with different rotational components is proposed. This strategy can be applied when the robot is constructed by a union of ellipsoids. Also, a procedure to iteratively refine the roadmap is introduced.

% The previous work \cite{yan2016path} only includes implementations for the 2D planning problems.
% , which extends the capabilities of the HRM planner for more complex robot models
% at each C-slice

\subsection{Overview of the Highway RoadMap planner} \label{sec:hrm-planner:overview}
\begin{figure}[!t]
\centering
\includegraphics[scale = 0.45, trim = 20 60 45 50, clip]{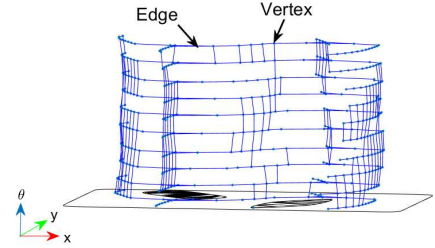}
\caption{The fully connected graph structure, generated from one simulation trial. The vertical axis represents the rotational angle; dots are vertices and line segments are edges.}
\label{fig:full_graph}
\end{figure}

The general workflow to construct this graph-based roadmap system is illustrated in Alg. \ref{algo:highway}. To visually demonstrate the concept, a fully connected graph obtained by running our algorithm in the planar case is shown in Fig. \ref{fig:full_graph}.

\begin{algorithm} [!t]
\SetAlgoLined
\SetKwInOut{Param}{Parameter}
\SetKwInOut{In}{Inputs}
\SetKwInOut{Out}{Outputs}

\In{$robot$: a union of ellipsoidal objects;\\
$obstacle$: a set of superquadric objects;\\
$arena$: a set of superquadric objects;\\
$endpts$: start and goal configurations}
\Param{$N_{\rm slice}$: number of C-slices;\\
$N_{\rm line}$: number of sweep lines;\\
$N_{\rm point}$: number of points for interpolation}
\Out{$roadmap$: a graph structure;\\
$path$: an ordered list of configurations}

$R$ $\leftarrow$ SampleOrientations($N_{\rm slice}$)\; \label{algo:highway:sample}
    
\ForEach{$i < N_{\rm slice}$}{
    $robot$.ForwardKinematics($R_i$) \; \label{algo:highway:fk}
	$roadmap \leftarrow$ ConstructOneSlice($robot$, $obstacle$, $arena$, $N_{\rm line}$) \; \label{algo:highway:one_slice}
}
    
\ForEach{$i < N_{\rm slice}$}{
	$roadmap \leftarrow$ ConnectAdjacentSlice($i$, $robot$, $R$, $N_{\rm point}$)\; \label{algo:highway:connect_slices}
}

$path$ $\leftarrow$ GraphSearch($roadmap$, $endpts$)\; \label{algo:highway:search}

\While{Not TerminationCondition}{
    $roadmap, path \leftarrow$ RefineExistRoadMap($robot$, $obstacle$, $arena$, $N_{\rm line}$, $R$) \; \label{algo:highway:refine_roadmap}
}
\caption{Highway RoadMap (HRM) Algorithm}
\label{algo:highway}
\end{algorithm}

The input of the $robot$ is a union of ellipsoids, including the body shapes and kinematic data. The kinematic data of each body part stores the relative rigid-body transformation with respect to the base. The input environmental data includes a set of superquadric objects that represent the $obstacles$ and $arena$. And the $endpts$ input indicates the start and goal configurations of the robot. There are two major input parameters: the number of C-slices $N_{\rm slice}$ and the initial number of sweep lines at each C-slice $N_{\rm line}$. These two parameters determine the initial resolution of the roadmap. $N_{\rm line}$ will be increased after the initial roadmap is built but the termination condition is not reached. The outputs of the algorithm are the $roadmap$ and $path$. The $roadmap$ is represented as a graph structure and the $path$ stores an ordered list of valid configurations from the start to the goal. This algorithm will terminate when any of the following conditions is satisfied: a valid path is found, the maximum planning time is reached or the maximum number of sweep lines is generated.

% can be reused to answer different path queries with the same environmental settings. T

In Alg. \ref{algo:highway}, Line \ref{algo:highway:sample} generates $N_{\rm slice}$ of discrete rotations in $\SO(d)$ and stored \textit{a priori}. Then, the forward kinematics is computed in Line \ref{algo:highway:fk} to rotate the rigid-body robot. At each fixed orientation, a subset of the C-space that only contains translation is built, denoted here as a ``C-slice'', in Line \ref{algo:highway:one_slice}. Once all C-slices are constructed, the vertices among adjacent C-slices are connected via a novel idea of ``bridge C-slice'' in Line \ref{algo:highway:connect_slices}. Each constructed C-slice only connects to its most adjacent C-slice. In Line \ref{algo:highway:search}, a graph search technique is applied to find a path from the starting configuration to the goal. In this work, $A^*$ algorithm \cite{hart1968formal} is used. When the termination condition is not satisfied, the roadmap is refined in an iterative way in Line \ref{algo:highway:refine_roadmap}.

%%%%%%%%%%%%%%%%%%%%%%%%%%%%%%%%
\subsection{Discretization of the robot orientations} \label{sec:hrm-planner:orientations}
Line \ref{algo:highway:sample} of Alg. \ref{algo:highway} pre-computes a set of orientation samples from $\SO(d)$. In the 2D case, uniformly distributed angles within the interval $[-\pi, \pi]$ are computed. In 3D, the icosahedral rotational symmetry group of the Platonic solid (consisting 60 elements) is used, which gives a finite and deterministic sampling of $\SO(3)$. The geodesic distances between two neighboring samples are almost uniformly distributed \cite{wulker2019quantizing}. Using this set of orientation samples, the rotational difference between two adjacent C-slices is smaller compared to non-uniform sample sets. Note that more rotations can be sampled to construct a denser roadmap per the user's choice, \ie using the strategies proposed in \cite{shoemake1992uniform,yershova2010generating}.

% More details on sampling $\SO(3)$ can be found in \cite{shoemake1992uniform,yershova2010generating}.

%%%%%%%%%%%%%%%%%%%%%%%%%%%%%%%%
\subsection{Construction of one C-slice} \label{sec:hrm-planner:one_slice}
The detailed procedure to construct one single C-slice (\ie Line \ref{algo:highway:one_slice} of Alg. \ref{algo:highway}) is outlined in Alg. \ref{algo:one_slice}. Within each C-slice, the closed-form Minkowski sum and difference for the bodies of robot are computed with the obstacles and arena, which results in $C_{\rm obstacle}$ and $C_{\rm arena}$, respectively (Line \ref{algo:one_slice:minkowski}). By sweeping parallel lines throughout the C-slice with a certain resolution (indicated by $N_{\rm line}$), the free portion of the C-slice ($C_{\rm free}$) is detected and represented as a set of line segments (Line \ref{algo:one_slice:sweep_line}). Furthermore, the middle point of each collision-free line segment is generated as the sampled vertices in the roadmap (Line \ref{algo:one_slice:generate_vertex}). Then, two vertices in adjacent sweep lines attempt to be connected by collision-free edges (Line \ref{algo:one_slice:connect_one_slice}).

\begin{algorithm} [!t]
\SetAlgoLined
\SetKwInOut{Return}{Return}

$C_{\rm obstacle}, C_{\rm arena} \leftarrow$ MinkowskiOperations($robot$, $obstacle$, $arena$)\; \label{algo:one_slice:minkowski}
$C_{\rm free} \leftarrow$ SweepLineProcess($C_{\rm obstacle}$, $C_{\rm arena}$, $N_{\rm line}$)\; \label{algo:one_slice:sweep_line}
$roadmap.vertex$.Append( GenerateVertex($C_{\rm free}$) )\; \label{algo:one_slice:generate_vertex}
$roadmap.edge$.Append( ConnectOneslice($C_{\rm free}$) )\; \label{algo:one_slice:connect_one_slice}
\Return{$roadmap$}

\caption{ConstructOneSlice($robot$, $obstacle$, $arena$, $N_{\rm line}$)}
\label{algo:one_slice}
\end{algorithm}

%%%%%%
\subsubsection{Minkowski operations for a multi-body robot} \label{sec:hrm-planner:one_slice:mink}
\begin{figure}[!t]
\centering
\subfloat[C-obstacle as the Minkowski sum boundaries of individual ellipsoidal bodies and their union.]{
	\label{fig:mink_union:c-obstacle}
	\centering
	\includegraphics[scale=0.4]{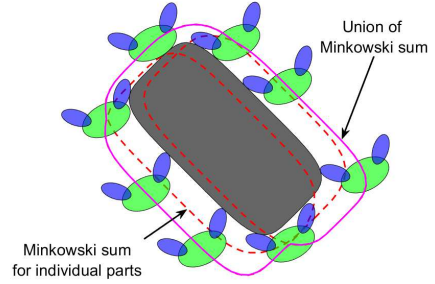}}
\hspace{0.1in}
\subfloat[Collision free C-space as an intersection of free space for individual robot parts.]{
	\label{fig:mink_union:c-space}
	\includegraphics[scale=0.4]{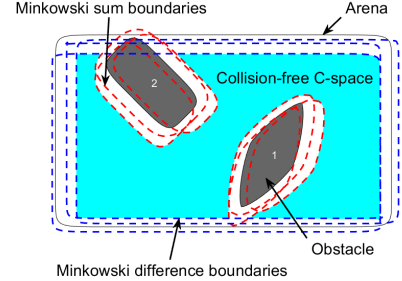}}
\caption{The characterization of the Minkowski sum between a convex superquadric and a union of ellipsoids.}
\label{fig:mink_union}
\end{figure}

At each C-slice, the closed-form Minkowski operations are computed to generate C-obstacles (\ie in Line \ref{algo:one_slice:minkowski} of Alg. \ref{algo:one_slice}). The robot is constructed by a finite union of $M$ rigidly connected ellipsoids $E_1, E_2, \dots, E_M$. Without loss of generality, $E_1$ is chosen as the base of the robot. The relative transformations between the base $E_1$ and other ellipsoidal parts $E_2, E_3, \dots, E_M$ are defined as $g_i = (R_i, {\bf t}_i)$ ($i = 2, \dots, M$), respectively. For a multi-link rigid-body robot, these relative transformations can be computed via forward kinematics with all the internal joints being fixed. With this definition and the property from Eq. \eqref{eq:math:union_mink}, the union of the Minkowski operations for all body parts can be expressed relative to one single reference point, which we choose as the center of the base ellipsoid $E_1$. In particular, for each ellipsoidal body $E_i$, a positional offset ${\bf t}_i$ is added to Eq. \eqref{eq:math:mink_bound_general_simplified}. For practical computational purposes, each Minkowski sum and difference boundary is discretized as a convex polygon in 2D and polyhedral mesh in 3D. The vertices of the discrete boundary are generated using the parametric expression of Minkowski operations. Figure \ref{fig:mink_union} shows the Minkowski sums of a multi-body robot at a fixed orientation (Fig. \ref{fig:mink_union:c-obstacle}) and the collision-free C-space in the corresponding C-slice (Fig. \ref{fig:mink_union:c-space}).

%%%%%%
\subsubsection{A sweep-line process to characterize free regions within one C-slice} \label{sec:hrm-planner:one_slice:sweep-line}
The general idea of the ``sweep-line'' process (\ie Line \ref{algo:one_slice:sweep_line} of Alg. \ref{algo:one_slice}) is analogous to raster scanning -- a set of parallel lines is defined to sweep throughout the whole C-slice. Theoretically, these parallel lines can be defined along any direction. But for simplicity of representation and storage, throughout this article, the lines are defined to be parallel to the basis axes of the coordinate system. Specifically, the sweep lines are parallel to $x$-axis and $z$-axis for the 2D and 3D case, respectively. Note that, in the 3D case, one could think the process as firstly sweeping planes through the 3D translational space, then sweeping lines within each plane. But in practice, there is no need to compute each plane completely that includes the silhouettes of C-obstacles. Instead, this work approximates each plane by a bundle of sweep lines, which are then used directly to compute free segments via line-obstacle intersections.

% Meanwhile, the sweep lines in different C-slices are generated with the same set of coordinates.

To generate collision-free configurations, segments on each sweep line within all C-obstacles and C-arenas are computed, denoted as $L_{\rm O}$ and $L_{\rm A}$, respectively. Then, the collision-free segments $L_{\rm free}$ can be computed as \cite{yan2016path,yan2014geometric}
\begin{equation}
L_{\rm free} = \bigcap_{i=1}^{M_A\times M} \, L_{\rm A_{i}} - \bigcup_{j=1}^{M_O\times M} \, L_{\rm O_{j}}
\end{equation}
where $M_A$ and $M_O$ are numbers of arenas and obstacles respectively. All $L_{\rm free}$ are stored in $C_{\rm free}$ (Line \ref{algo:one_slice:sweep_line} of Alg. \ref{algo:one_slice}). Then, collision-free vertices are generated as the middle point of each $L_{\rm free}$ (Line \ref{algo:one_slice:generate_vertex} of Alg. \ref{algo:one_slice}). Afterwards, more vertices can be generated as an enhancement step. An example is given and applied throughout this article. Denote $L_{j,k}$ as the $k^{\rm th}$ free segment of $j^{\rm th}$ sweep line, with $V_{j,k}$ being its corresponding middle point. Firstly, $L_{j+1,k_2}$ is projected onto $L_{j,k_1}$. If the projection overlaps with $L_{j,k_1}$ but $V_{j, k_1}$ is not within the overlapping segment, a new vertex within the overlapping segment that is nearest to $V_{j, k_1}$ is added to the vertex list. The resulting new vertex is closer to $V_{j+1,k_2}$ than $V_{j,k_1}$ does, which gives higher chance to make the further connection success, especially in narrow regions.

\begin{figure} [!t]
\centering
\includegraphics[scale = 0.45, trim = 50 30 30 30, clip]{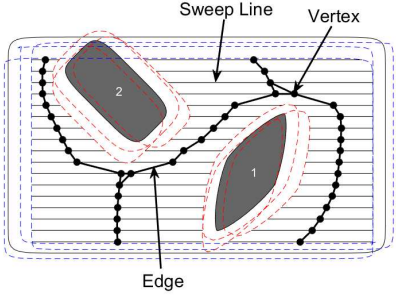}
\caption{The sweep line process for detecting free space and construct sub-graph in one C-slice.}
\label{fig:one_slice}
\end{figure}

Once a list of collision-free vertices is generated, the next step is to connect them (Line \ref{algo:one_slice:connect_one_slice} of Alg. \ref{algo:one_slice}). In this work, only two vertices in adjacent sweep lines attempt to be connected with a straight line segment. Assume a candidate connection is attempted between $V_{j, k_1}$ and $V_{j+1, k_2}$. The connection validity is checked by computing the intersections between the line segment $\overline{V_{j, k_1} V_{j+1, k_2}}$ and all meshed C-obstacles. If the segment is outside all C-obstacles, the whole edge is guaranteed to be collision-free. Figure \ref{fig:one_slice} shows the decomposed C-space in one slice of a planar case. The horizontal raster lines indicate the collision-free line segments. This method provides a continuous way of validating edges within each C-slice, in the sense that the whole edge is checked without interpolation.

%%%%%%%%%%%%%%%%%%%%%%%%%%%%
\subsection{Vertex connections between adjacent C-slices} \label{sec:hrm-planner:slice-connection}
Since each C-slice only represents one orientation of the robot, rotational motions are required when connecting different C-slices. A novel ``bridge C-slice'' method is proposed (\ie in Line \ref{algo:highway:connect_slices} of Alg. \ref{algo:highway}) to guarantee that the vertices at different C-slices can be safely connected without performing explicit collision detection. Algorithm \ref{algo:connect_slices} outlines this new local planner. The general idea is to construct a new C-slice based on an enlarged ellipsoidal void that encloses the robot at two configurations and compute a translational sweep volume that bounds the whole transition.

% The previous work showed the idea of constructing a ``local C-space'' by enlarging the ellipsoidal robot, and connecting two poses inside a convex polyhedral C-space \cite{ruan2018path}. The method assumes ``small motion'' between two configurations. Alternatively, in this article,

\begin{algorithm} [!t]
\SetAlgoLined
\SetKwInOut{Return}{Return}

$R_{\rm near} \leftarrow$ GetAdjacentSlice($i$, $R$)\; \label{algo:connect_slices:adj_slice}
TFE $\leftarrow$ ComputeTightlyFittedEllipsoids($robot$, $R_{i}$, $R_{\rm near}$, $N_{\rm point}$) \; \label{algo:connect_slices:tfe}
$C_{\rm obstacle}, C_{\rm arena} \leftarrow$ MinkowskiOperations(TFE, $obstacle$, $arena$)\; \label{algo:connect_slices:minkowski}

\ForEach{{\rm vertex $V_1$ in current C-slice}}{
    $\{ V_{1, \rm near} \} \leftarrow$ NeighborVerticesInAdjacentSlice( $roadmap.vertex$ ) \; \label{algo:connect_slices:nn}
    \ForEach{$V_2 \in \{ V_{1, \rm near} \}$}{
        $\{ V_{\rm step} \} \leftarrow$ PathInterp($V_1$, $V_2$, $N_{\rm point}$) \; \label{algo:connect_slices:interp}
        \If{ {\rm IsPathValid($\{ {\bf t}_{\rm step} \}$, $C_{\rm obstacle}$, $C_{\rm arena}$)} \label{algo:connect_slices:validation} }{
            $roadmap.edge$.Append($\{ V_1, V_2 \}$) \;
        }
    }
}
\Return{$roadmap$}

\caption{ConnectAdjacentSlice($i$, $robot$, $R$, $N_{\rm point}$)}
\label{algo:connect_slices}
\end{algorithm}

\subsubsection{General ideas of the ``bridge C-slice'' local planner}
Each C-slice only attempts to connect with one adjacent C-slice, which is searched at the beginning in Line \ref{algo:connect_slices:adj_slice} of Alg. \ref{algo:connect_slices}. And the metric that evaluates adjacency is based on the distance of the rotational components. In the 3D case, for instance, the Euclidean distance between the quaternion parameterization of the two bodies is used. The core steps in Alg. \ref{algo:connect_slices} are Line \ref{algo:connect_slices:tfe}, which constructs an enlarged tightly-fitted ellipsoid (TFE) for each robot part, and Line \ref{algo:connect_slices:validation}, which validates the whole edge connecting the two vertices. 

Suppose that the robot is moving from vertex $V_1 = (R_1, {\bf t}_1)$ to $V_2 = (R_2, {\bf t}_2)$, where $R_i$ and ${\bf t}_i \, (i=1,2)$ represents the rotation and translation part of vertex $V_i$, respectively. The idea here is to enclose the motions of each ellipsoidal part of the robot, \ie $E_k$, between the two configurations by a tightly fitted sweep volume, which is guaranteed to be collision-free. The intermediate configurations between $V_1$ and $V_2$ can be computed using interpolation technique. To construct the sweep volume, a tightly-fitted concentric ellipsoid (TFE) for each $E_k$ at all orientations from the interpolated motions is computed, which will be detailed in Sec. \ref{sec:hrm-planner:slice-connection:tfe}. The computed TFE is the void that guards the safe motions of the actual ellipsoidal part. Then, the computed TFE translates from ${\bf t}_1$ to ${\bf t}_2$ following the interpolated path (\ie $\{ {\bf t}_{\rm step} \}$) of $E_k$'s center. The resulting sweep volume bounds the whole transition of $E_k$ between the two configurations. To ensure that each computed TFE stays inside the collision-free space, one could query the inside-outside status of all the intermediate translation parts $\{ {\bf t}_{\rm step} \}$ with all C-obstacles and C-arena. Then, if all the positions from $\{ {\bf t}_{\rm step} \}$ are valid, the sweep volume is guaranteed to be safe. Therefore, the whole transition for the ellipsoidal part $E_k$ is collision-free.

\begin{figure}[!t]
\centering
\subfloat[Sweep volume for individual elliptical part.]{
\label{fig:sweep_volume:individual}
\centering
\includegraphics[scale=0.4]{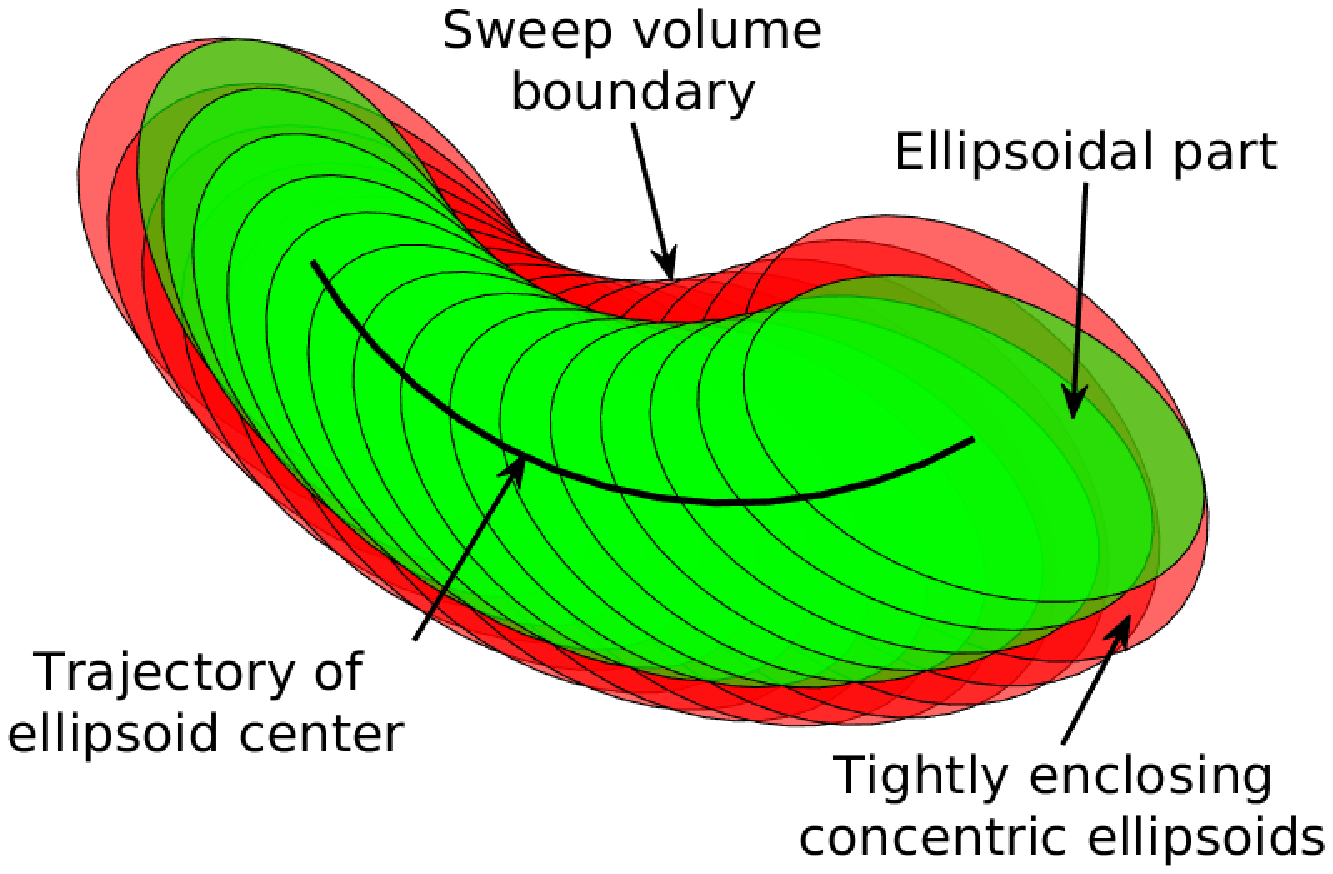}}
\hspace{0.1in}
\subfloat[Sweep volume for the whole multi-body robot.]{
\label{fig:sweep_volume:whole}
\centering
\includegraphics[scale=0.4]{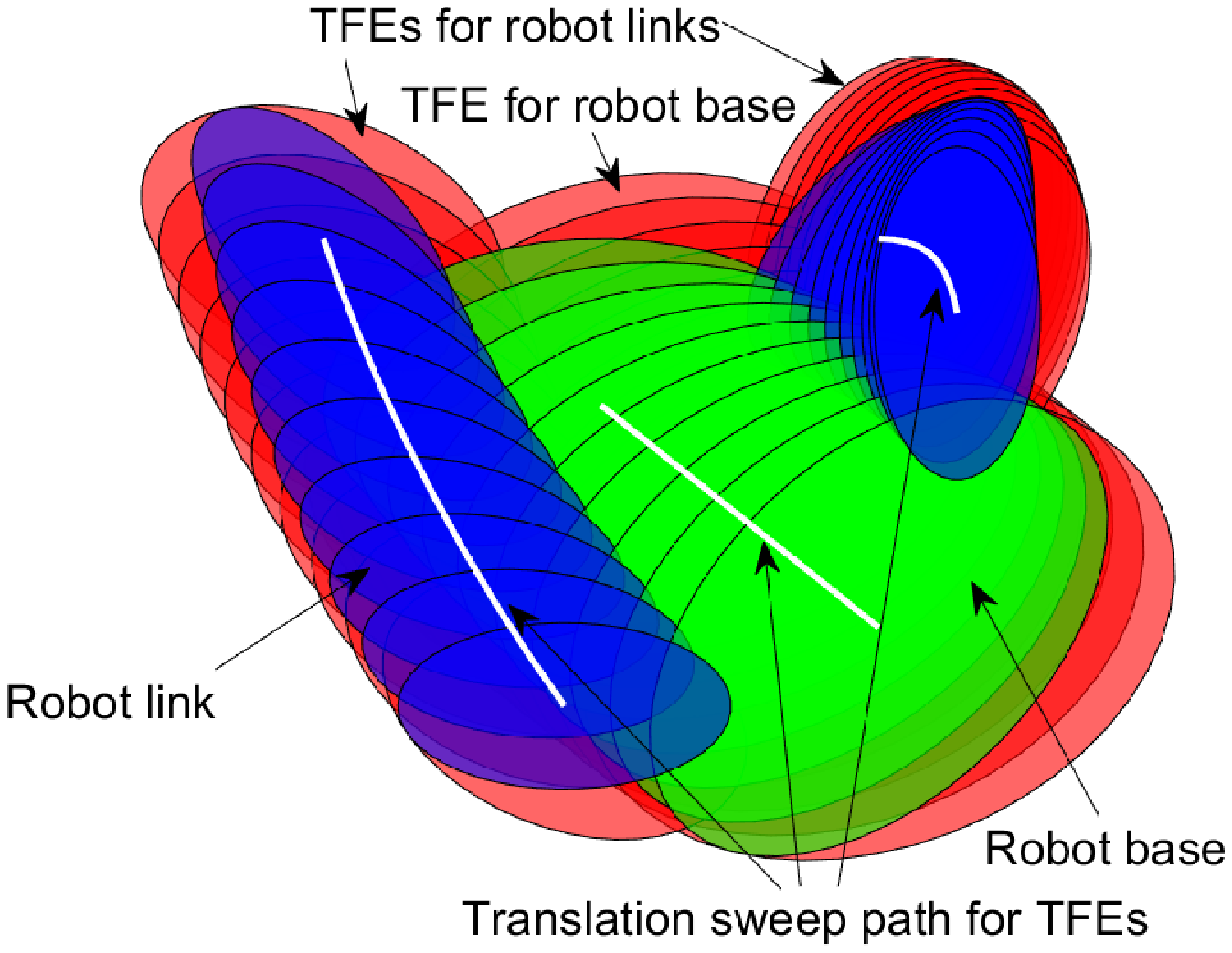}}
\caption{2D example illustrating the sweep volume idea based on the sliding of tightly-fitted ellipsoids.}
\label{fig:sweep_volume}
\end{figure}

Figure \ref{fig:sweep_volume:individual} shows the procedure of constructing the sweep volume for an individual body part. And Fig. \ref{fig:sweep_volume:whole} illustrates the union of sweep volumes that encloses the whole multi-body robot in the planar case. The robot base follows a 2D straight line with rotations, and the TFEs of different body parts follow different paths (as show in white curves). In this process, the TFE for each body part translates with respect to its own center individually. This differs from the operations within one C-slice, which requires an offset to the C-obstacle and C-arena boundaries in order to make the robot as a whole rigid body. The reason is that, as Fig. \ref{fig:sweep_volume:whole} shows, the motion of each robot part is no longer a pure translation. Therefore, the reference points of Minkowski operations for different body parts have different trajectories to follow. The transition for the whole robot is guaranteed safe if all the individual reference points are within their own free space. 

%%%%%%
\subsubsection{Computational procedure for ``Tightly-Fitted Ellipsoids''} \label{sec:hrm-planner:slice-connection:tfe}
Line \ref{algo:connect_slices:tfe} of Alg. \ref{algo:connect_slices} generates the TFE for each individual part of the robot. The detailed computational procedure is shown in Alg. \ref{algo:tfe}. Firstly, an interpolation of the orientations is computed (Line \ref{algo:tfe:interp}). The number of intermediate orientations is pre-defined by users as the parameter $N_{\rm point}$. Then, the TFE set, represented as a set of ellipsoids, is initiated as the robot at $i^{\text{th}}$ orientation (Line \ref{algo:tfe:init}). For each interpolated step, the orientations of all the robot parts are updated (Line \ref{algo:tfe:update_rot}). Finally, the updated TFE for each robot part is generated by computing the minimum volume concentric ellipsoid (MVCE) (introduced in Sec. \ref{sec:math:mvce}) of the current TFE and each ellipsoidal part at the new orientation (Line \ref{algo:tfe:mvce_step}). This procedure requires $N_{\rm point}$ iterations so that all the interpolated orientations between two C-slices can be fully encapsulated. 

\begin{algorithm} [!t]
\SetAlgoLined
\SetKwInOut{Return}{Return}

$\{ R_{\rm step} \} \leftarrow$ RotationInterpolation($R_i$, $R_j$, $N_{\rm point}$) \; \label{algo:tfe:interp}
TFE $\leftarrow$ $robot$.ForwardKinematics($R_i$) \; \label{algo:tfe:init}
\ForEach{$R_{\rm step}$}{
$robot.$ForwardKinematics($R_{\rm step}$) \; \label{algo:tfe:update_rot}
    \ForEach{$robot$ {\rm part} $E_k$}{
    $\text{TFE}[k]$ $\leftarrow$ MVCE($\text{TFE}[k]$, $E_k$) \; \label{algo:tfe:mvce_step}
    }
}
\Return{TFE (a set of TFEs for different robot parts)}

\caption{ComputeTightlyFittedEllipsoids($robot$, $R_i$, $R_j$, $N_{\rm point}$)}
\label{algo:tfe}
\end{algorithm}

%%%%%%
\subsubsection{Vertex connections based on bridge C-slice calculations}
A ``bridge C-slice'' is constructed via closed-form Minkowski operations between the computed TFE and obstacles/arena (Line \ref{algo:connect_slices:minkowski} of Alg. \ref{algo:connect_slices}). Then, the algorithm attempts to connect all the existing vertices to their nearest neighbors within the adjacent C-slice. The nearest neighbors of a vertex are defined as located within the same sweep line (Line \ref{algo:connect_slices:nn} of Alg. \ref{algo:connect_slices}). 

For each candidate connection, the robot is transformed according to the interpolated configurations between two vertices (Line \ref{algo:connect_slices:interp} of Alg. \ref{algo:connect_slices}). Note that the rotation part of each interpolated motion needs to match those when computing TFEs (\ie Line \ref{algo:tfe:interp} of Alg. \ref{algo:tfe}). This is not hard to achieve for a typical interpolation of rigid-body motions, even when a simultaneous rotation and translation is considered. For example, this article uses interpolations in $\SE(3)$ of the form $ g_{\rm step} = g_1 \exp[ \tau \log (g^{-1}_1 \, g_2) ] $, where $\tau \in [0,1]$ parameterizes the transition, $g_1 \,, g_2 \in \SE(3)$ are the two end points of the interpolation, $\exp[\cdot]$ and $\log(\cdot)$ are matrix exponential and logarithm respectively. The rotation part of each step is the same with interpolations on $\SO(3)$ since the group operation for the rotation part is not affected by the translation part.

With the C-obstacle and C-arena being computed for the TFE of each individual robot part, the next step is to check the validity of the translation motions of each TFE (Line \ref{algo:connect_slices:validation} of Alg. \ref{algo:connect_slices}). The inside-outside status of this point with all the C-obstacles are queried. If any of the center point is inside any C-obstacle, the validating process is terminated and the corresponding connection is discarded. Otherwise, further checks for other ellipsoidal parts are conducted until all the parts are checked. 

The sweep volume gives a conservative encapsulation of the robot transitions between two vertices. But if the orientation samplings are incremental and uniform, there will not be a large rotational difference between adjacent C-slices. Thus, the extra free space inside the sweep volume will be small.

%%%%%%%%%%%%%%%%%%%%%%%%%%%%
\subsection{Refinement of the existing roadmap} \label{sec:hrm-planner:refine-exist}

Line \ref{algo:highway:refine_roadmap} of Alg. \ref{algo:highway} tries to refine the existing roadmap by increasing the density of sweep lines at each existing C-slice. This process will be triggered when the termination conditions are not satisfied after building and searching the whole roadmap. Detailed procedures are presented in Alg. \ref{algo:refine_roadmap}. Firstly, $N_{\rm line}$ is doubled (Line \ref{algo:refine_roadmap:double_param}). Then, at each C-slice, the same procedure with Alg. \ref{algo:one_slice} that constructs one C-slice with more sweep lines is performed (Line \ref{algo:refine_roadmap:one_slice}). Note that the C-obstacles are stored during the initial construction of C-slices, then they can be directly retrieved without re-computing. Afterwards, the new denser set of vertices attempts to connect with the old vertices within the same C-slice (Line \ref{algo:refine_roadmap:connect_existing}). This process firstly locates the vertices that have the same rotation part in the existing roadmap. Then, each new vertex attempts to connect with nearby existing vertex using the same procedure in Line \ref{algo:one_slice:connect_one_slice} of Alg. \ref{algo:one_slice}. Once connections are done, the graph search is performed again (Line \ref{algo:refine_roadmap:search}).

\begin{algorithm} [!t]
\SetAlgoLined
\SetKwInOut{Return}{Return}

$N_{\rm line} = N_{\rm line} * 2$\; \label{algo:refine_roadmap:double_param}
\ForEach{$R_i \in R$}{
    $robot$.ForwardKinematics($R_i$)\;
    $roadmap \leftarrow$ ConstructOneSlice($robot$, $obstacle$, $arena$, $N_{\rm line}$)\; \label{algo:refine_roadmap:one_slice}
    $roadmap \leftarrow$ ConnectExistSlice($i$, $R$)\; \label{algo:refine_roadmap:connect_existing}
    $path$ $\leftarrow$ GraphSearch($roadmap$, $endpts$)\; \label{algo:refine_roadmap:search}
    
    \If{TerminationCondition}{
        break\;
    }
}
\Return{$roadmap$, $path$}

\caption{RefineExistRoadMap($robot$, $obstacle$, $arena$, $N_{\rm line}$, $R$)}
\label{algo:refine_roadmap}
\end{algorithm}

% Probabilistic Highway Roadmap planner
\section{Hybrid Probabilistic Variation of Highway RoadMap Planner for Articulated Robots with Ellipsoidal Components}
\label{sec:prob-hrm-planner}
The original HRM planner in Sec. \ref{sec:hrm-planner} only designs for the case when robot parts are rigidly connected to each other. This limits its ability to extend to higher dimensional configuration space, \ie $\SE(d) \times (S^1)^n$. To avoid the exponential computational complexity in concatenating all possible combinations of the base pose and joint angles, a hybrid algorithm is proposed here. The general idea is to combine with sampling-based planners, which are proved to be advantageous in dealing with the ``curse of dimensionality''. Algorithm \ref{algo:prob-highway} shows the general workflow of the proposed hybrid probabilistic Highway RoadMap (Prob-HRM) planner. Prob-HRM mainly differs from the original HRM algorithm in that it utilizes the idea of random sampling for rotational components of the robot, \ie the orientation of the base part and all joint angles. 

\begin{algorithm} [!t]
\SetAlgoLined
\SetKwInOut{Param}{Parameter}
\SetKwInOut{In}{Inputs}
\SetKwInOut{Out}{Outputs}

\In{$robot$: a list of ellipsoidal objects and robot kinematic information;\\
$obstacle$: a set of superquadric objects;\\
$arena$: a set of superquadric objects;\\
$endpts$: start and goal configurations; }
\Param{$N_{\rm line}$: number of sweep lines;\\
$N_{\rm point}$: number of points for interpolation}
\Out{$roadmap$: a graph structure;\\
$path$: an ordered list of configurations}

$N_{\rm slice} = 0$ \;
\While{Not TerminationCondition}{
    $R_{\rm current} \leftarrow$ RandomSampleRobotShape()\; \label{algo:prob-highway:sampling}
    $robot$.ForwardKinematics($R_{\rm current}$)\; \label{algo:prob-highway:fk}
    $roadmap \leftarrow$ ConstructOneSlice($robot$, $obstacle$, $arena$, $N_{\rm line}$)\; \label{algo:prob-highway:one_slice}
    $roadmap \leftarrow$ ConnectAdjacentSlice($robot$, $R_{\rm current}$, $N_{\rm point}$)\; \label{algo:prob-highway:connect_slices}
    }
    $path$ $\leftarrow$ GraphSearch($roadmap$, $endpts$)\; \label{algo:prob-highway:search}
    
    $R$.Append( $R_{\rm current}$ )\;
    $N_{\rm slice} = N_{\rm slice} + 1$ \;

    \If{Refine current $roadmap$}{
        $roadmap, path \leftarrow$ RefineExistRoadMap($robot$, $obstacle$, $arena$, $N_{\rm line}$, $R$)\; \label{algo:prob-highway:refine}
    }
\caption{Probabilistic Highway RoadMap (Prob-HRM) Algorithm}
\label{algo:prob-highway}
\end{algorithm}

% In contrast, HRM deterministicly generates rotations of the robot with fixed resolution during the run time. 

The robot with fixed rotational components is called a ``shape'' \cite{lien2008hybrid}, and a single C-slice is computed for each robot shape. Since for each shape, the internal joint angles are fixed, computations within the same C-slice in Prob-HRM stay the same with those in HRM, \ie Line \ref{algo:prob-highway:one_slice} of Alg. \ref{algo:prob-highway} are the same with the corresponding subroutine in Alg. \ref{algo:highway}. Other subroutines are also easy to be ported from the original HRM to Prob-HRM. In particular, the only difference for vertex connections among adjacent C-slices (Line \ref{algo:prob-highway:connect_slices}) with that in HRM is that the connection attempts are made only for the new C-slice in the current loop. In HRM, as a comparison, the adjacent C-slices are connected in the end after all C-slices are generated. Also, the graph search process is conducted each time after the new C-slice is connected to the graph (as in Line \ref{algo:prob-highway:search}). In contrast, for HRM, the graph search is conducted once after the whole graph is built. The new subroutines in Prob-HRM are the random sampling of robot shapes (Line \ref{algo:prob-highway:sampling}) and the computations of forward kinematics (Line \ref{algo:prob-highway:fk}) in each loop. To sample a shape, the orientation of the robot base is uniformly and randomly sampled \cite{shoemake1992uniform}, followed by random sampling of joint angles within their ranges. After that, the forward kinematics is computed to get the poses of all the robot body parts with respect to the world frame. When $N_{\rm slice}$ reaches a certain number but the termination conditions are still not satisfied, the C-slice exploration is paused and the refinement of the current roadmap is triggered. In practice, this refinement process is triggered when each 60 new C-slices are generated. The refinement procedure in Line \ref{algo:prob-highway:refine} is the same with Alg. \ref{algo:refine_roadmap}. Once all existing C-slices are refined but the algorithm is still not terminated, new C-slices exploration will be resumed. Note that $N_{\rm slice}$ in HRM is no longer a pre-defined parameter of Prob-HRM since the orientation of the robot base and joint angles are updated online.

% Benchmark Simulations
\section{Benchmarks on Path Planning for Ellipsoidal Robots in Superquadric Environments} 
\label{sec:benchmark}
This section compares the performance of the proposed HRM and Prob-HRM planners with some well-known sampling-based motion planners. The proposed planners are written in C++. The baseline sampling-based planners to be compared are sourced from the ``Open Motion Planning Library (OMPL)'' \cite{sucan2012open}. All the benchmarks are conducted on Ubuntu 16.04 using an Intel Core i7 CPU at $3.40 \text{ GHz} \times 8$. 

%  The configuration spaces studied include $\SE(3)$ and $\SE(3) \times (S^1)^n$. The robots are enclosed by a union of ellipsoids and the obstacles are modeled as superquadrics.

%%%%%%%%%%%%%%%%%%%%%%%%%%%%%%%%%%%%%%%%%%%%%%%%%%%%%%%%%%%%%
\subsection{Planning environment and robot type settings}
\begin{figure*} [!t]
	\centering
	\subfloat[3D maze map, rabbit-shape robot]{
		\label{fig:bench:map:maze_3d_rabbit}
		\centering
		\includegraphics[scale=0.5, trim = 60 60 50 60, clip]{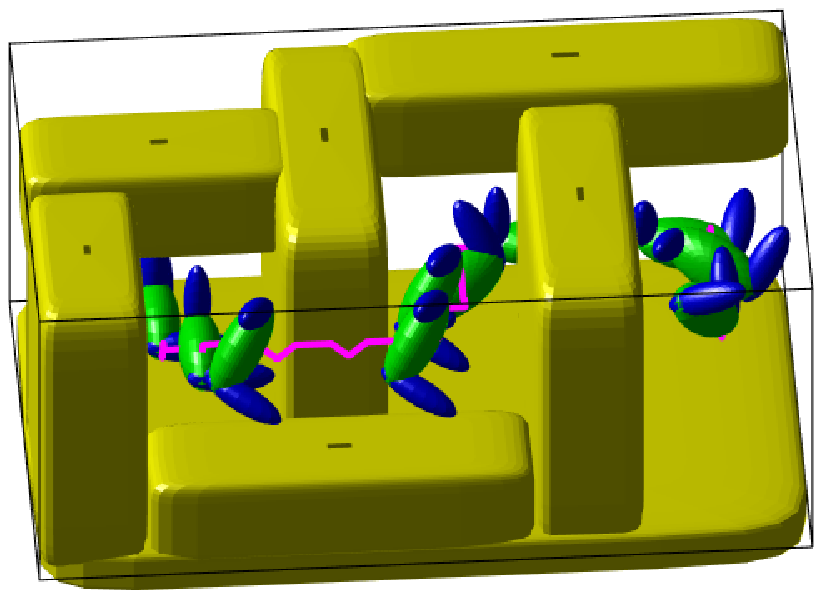}}
	\subfloat[3D home map, chair object]{
		\label{fig:bench:map:home_3d_chair}
		\centering
		\includegraphics[scale=0.5, trim = 80 60 60 50, clip]{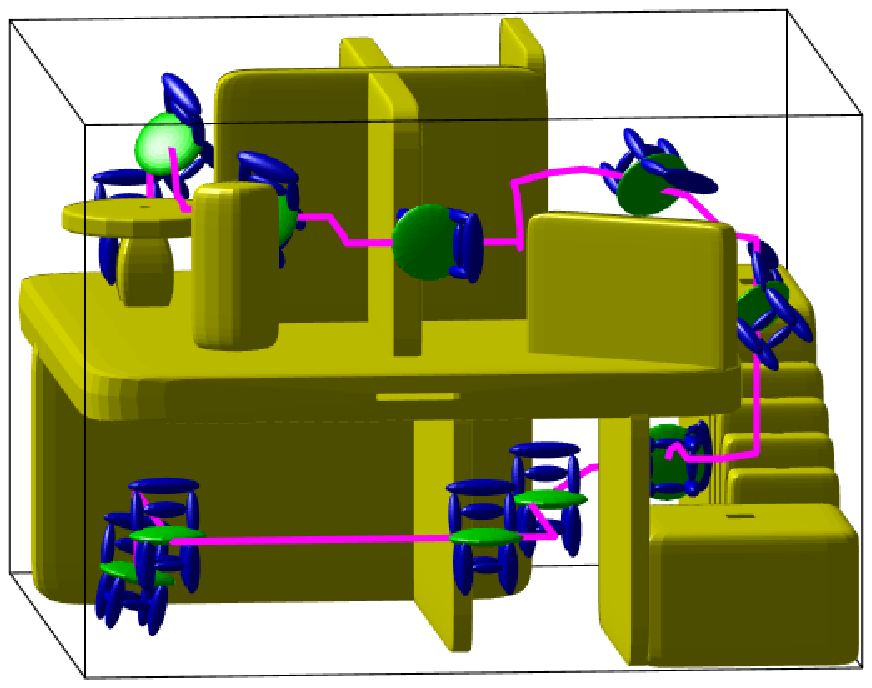}}
	\subfloat[3D cluttered map, snake-like robot]{
		\label{fig:bench:map:cluttered_3d_snake}
		\centering
		\includegraphics[scale=0.5, trim = 60 60 40 60, clip]{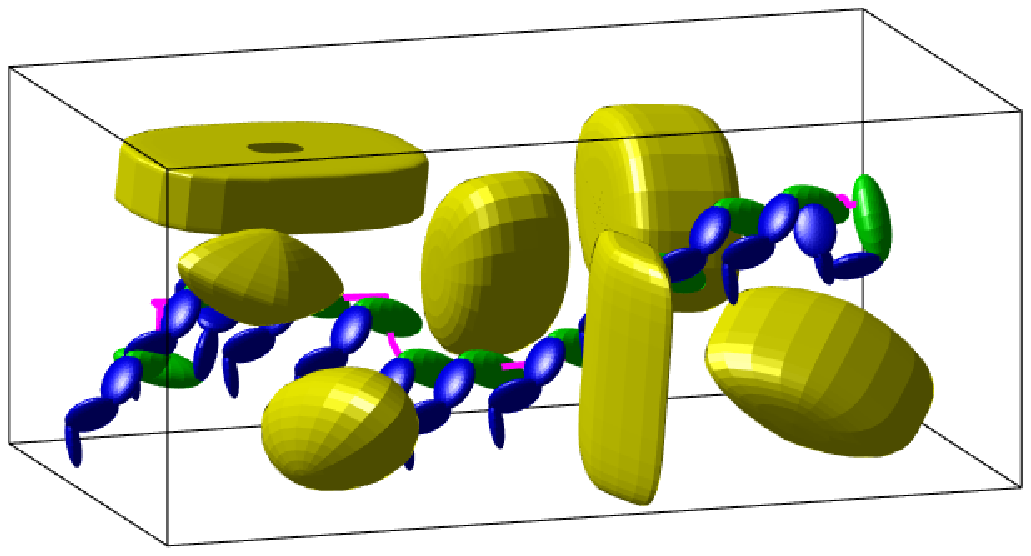}}
	\hspace{0.1in}
	\subfloat[3D narrow window map, snake-like robot]{
	    \label{fig:bench:map:narrow_3d_snake}
	    \centering
	    \includegraphics[scale=0.5, trim = 60 60 50 50, clip]{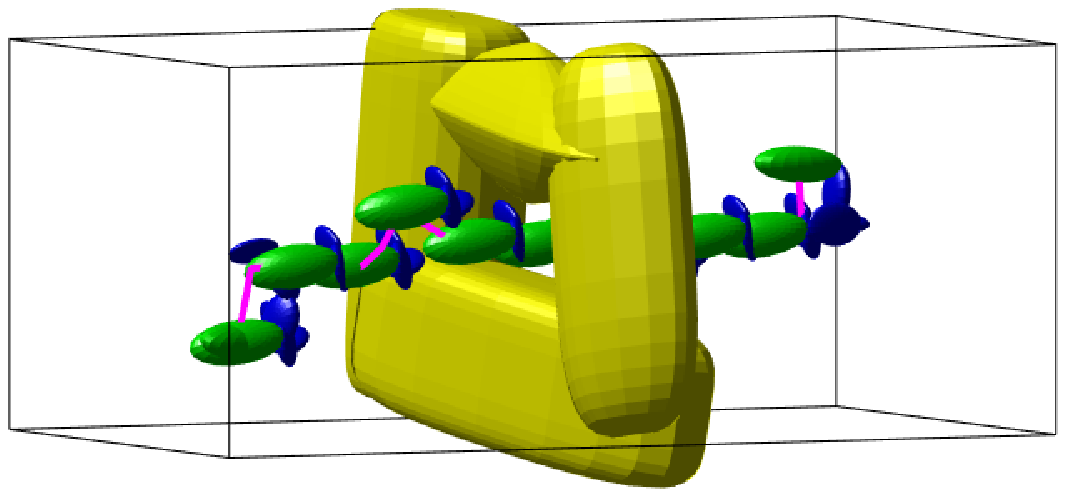}}
	\subfloat[3D sparse map, tree-like robot]{
		\label{fig:bench:map:sparse_3d_tree}
		\centering
		\includegraphics[scale=0.5, trim = 60 60 60 60, clip]{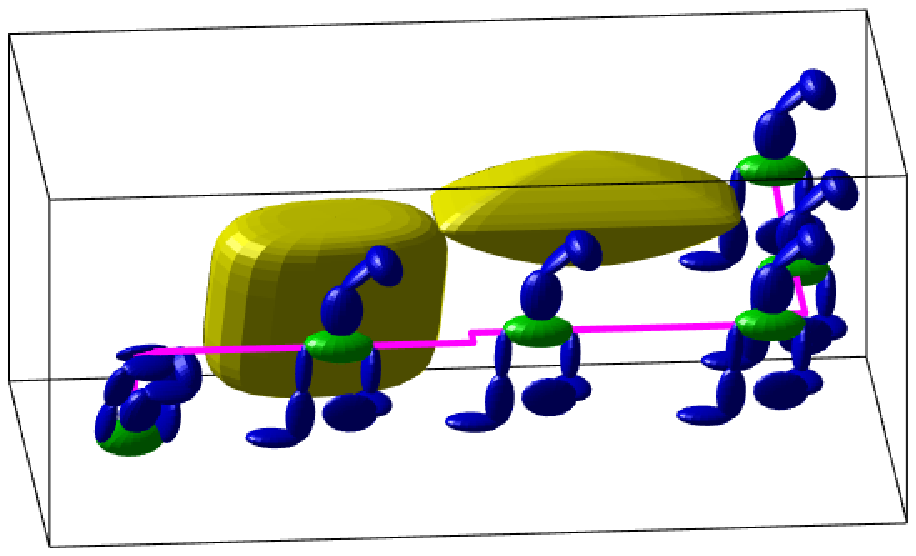}}
		
	\caption{Demonstration of path planning solutions using our proposed HRM-based planners for different types of robots in different environments. Problems with rigid-body and articulated robots are planned using HRM and Prob-HRM respectively. Obstacles are 3D superquadrics and the robots are constructed by unions of 3D ellipsoids. The magenta curve represents the solved path of the robot base center that is projected from C-space to Euclidean space.}
	\label{fig:bench:map}
\end{figure*}

Figure \ref{fig:bench:map} shows the planning environments and the solved paths for different robots using our proposed HRM or Prob-HRM planners. Both rigid-body and articulated robots are considered. \\
The rigid-body robots include: 
\begin{itemize}
\item tilted rabbit (Fig. \ref{fig:bench:map:maze_3d_rabbit}), with 3 body parts being rigidly and serially connected but not co-planar; and
\item rigid object with 13 parts, resembling a common chair (Fig. \ref{fig:bench:map:home_3d_chair}).
\end{itemize}
The articulated robots include:
\begin{itemize}
\item snake-like robot (Figs. \ref{fig:bench:map:cluttered_3d_snake} and \ref{fig:bench:map:narrow_3d_snake}) which is serially configured with one movable base and 3 links (totally 9 degrees of freedom); and
\item tree-like robot (Fig. \ref{fig:bench:map:sparse_3d_tree}), which is a tree structure with one movable base in the middle and 3 branches of RRR-typed serial linkages (totally 15 degrees of freedom).
\end{itemize}
The planning environments being considered include: 
\begin{itemize}
\item spatial maze map (Fig. \ref{fig:bench:map:maze_3d_rabbit}) with more narrow corridors;
\item home map (Fig. \ref{fig:bench:map:home_3d_chair}) that is constructed as a 2-floor house with walls, corridors, stairs and tables;
\item cluttered map (Fig. \ref{fig:bench:map:cluttered_3d_snake}) with obstacles in arbitrary poses;
\item narrow window map (Fig. \ref{fig:bench:map:narrow_3d_snake}) that includes one wall with a small window available for the robot to move through;
\item sparse map (Fig. \ref{fig:bench:map:sparse_3d_tree}) with only two obstacles.
\end{itemize}

% The maps and robots used for benchmarks are summarized in this subsection. 

%%%%%%%%%%%%%%%%%%%%%%%%%%%%%%%%%%%%%%%%%%%%%%%%%%%%%%%%%%%%%
\subsection{Parameter settings for planners}
The compared baseline sampling-based planners from OMPL are PRM \cite{kavraki1996probabilistic}, Lazy PRM \cite{bohlin2000path}, RRT \cite{lavalle1998rapidly}, RRT Connect \cite{kuffner2000rrt} and EST \cite{hsu1997path}. Moreover, different sampling methods for PRM are also considered, including uniform random sampling (Uniform), obstacle-based sampling (OB) \cite{rodriguez2006obstacle}, Gaussian sampling (Gaussian) \cite{boor1999gaussian}, bridge test (Bridge) \cite{hsu2003bridge} and maximized clearance sampling (MC). We conduct 50 planning trials per planner per map. A time limit of 300 seconds is set for one planning trial for all planners. A planning trial is considered failure if the time exceeds this limit. 

% The following describes the parameters used for both HRM-based and sampling-based planners.

%%%%%%
\subsubsection{Parameters for our proposed HRM-based planners}
Table \ref{tab:param_high} shows the parameters of the HRM-based planners for each scenarios in Fig. \ref{fig:bench:map}. $N_{\rm slice}$ is only defined for HRM planner, as explained in Sec. \ref{sec:hrm-planner:orientations}. For scenes including articulated robot (\ie snake and tree), $N_{\rm slice}$ is not a pre-defined parameter. The initial value of $N_{\rm line}$ is a parameter defined by user or computed according to the planning scenario. In the following benchmark studies, the latter case is used. Based on the sizes of the obstacles and robot parts, the initial number of lines along a certain direction (\ie $N_{\rm dir}$) is computed by
\begin{equation}
N_{\rm dir} = \left\lfloor \frac{{\bf a}_{\rm dir}(A) - \max_{i} {\bf a}(E_i)}{\min_{j} {\bf a}(O_j)} \right\rfloor \,,
\label{eq:bench:num_line}
\end{equation}
where ${\bf a}_{\rm dir}(A)$, ${\bf a}(E_i)$ and ${\bf a}(O_j)$ denote the semi-axis length of the arena $A$ along direction ${\rm dir}$, an ellipsoidal robot part $E_i$ and an obstacle $O_j$, respectively. In 3D case, $N_{\rm line}$ is a multiplication of the numbers of lines along $x$ and $y$ axes directions, \ie $N_{\rm line} (N_x \times N_y)$.

% the number of which for all the maps is 60 -- the number of discrete rotations in $\SO(3)$ (

\begin{table} [!t]
\centering
\caption{Parameters for HRM-based planners in scenarios from Fig. \ref{fig:bench:map}}
\label{tab:param_high}
	
\begin{tabular}{cccc}
	\hline
	Map & Robot & $N_{\rm slice}$ & $N_{\rm line} \, (N_x \times N_y)$ \\ \hline
	Maze & Rabbit & 60 & 55 ($11 \times 5$) \\
	Home & Chair & 60 & 400 ($20 \times 20$) \\
	Cluttered & Snake & -- & 72 ($12 \times 6$) \\
	Narrow & Snake & -- & 18 ($6 \times 3$) \\
	Sparse & Tree & -- & 10 ($5 \times 2$) \\
	\hline
\end{tabular}
\end{table}

%%%%%%
\subsubsection{Parameters for sampling-based planners} \label{sec:benchmark:parameter:sampling-based}
For sampling-based planners, the choice of a relatively fast collision checker is essential. We choose the open-source and widely-used library, \ie ``Flexible Collision Library (FCL)'' \cite{pan2012fcl}, as an external plug-in for collision detection between robot parts and obstacles. In particular, a special and efficient collision object from FCL is applied for ellipsoidal parts of the robot. The library uses 12 extreme vertices to outer bound the exact ellipsoidal surface, resultng in a discretized polyhedral model. For superquadrics, their surfaces are discretized as triangular meshes based on the parametric expressions. Then, the bodies can be seen as convex polyhedra. The collision objects are generated \textit{a priori}, and the collision queries are made online by only changing the poses of each body part. 

\begin{figure} [!t]
	\centering
	\includegraphics[scale=0.45]{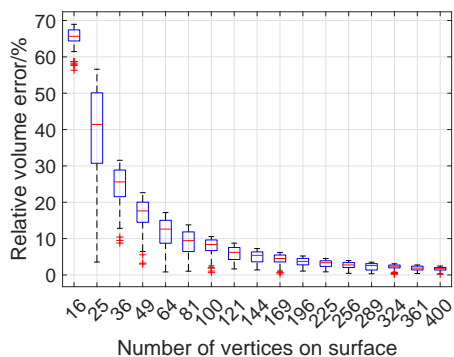}
	\caption{Relative volume for discretized superquadrics.}
	\label{fig:rel_vol_sq}
\end{figure}

Since the efficiency and accuracy of collision checking highly depend on the quality of discretization, we provide a statistical evaluation to determine the number of vertices for the discrete superquadric surface. The evaluation metric is based on the relative volume difference between the ground truth and fitted geometries, \ie
\begin{equation}
\kappa_{\rm volume} = \frac{\left| \rm Vol_{fitted} - Vol_{true} \right|}{\rm Vol_{true}} \times 100 \% \,,
\label{eq:rel_vol_sq}
\end{equation}
where, ${\rm Vol_{true}}$ and ${\rm Vol_{fitted}}$ denote the volume of the ground truth and fitted geometries, respectively. Here the ground truth is considered as the superquadric and the fitted object is the convex polyhedron. The volume of a superquadric body can be computed as $ {\rm Vol_{SQ}} = 2 abc \epsilon_1 \epsilon_2 \beta(\frac{\epsilon_1}{2}+1, \epsilon_1) \beta(\frac{\epsilon_2}{2}, \frac{\epsilon_2}{2}) $, where $\beta(x,y) = 2 \int_{0}^{\pi/2} \sin^{2x-1}\phi \cos^{2y-1}\phi \,\, d\phi$ is the beta function. $\kappa_{\rm volume}$ are computed for different numbers of vertices on the superquadric surface. For each discretization, 100 random superquadric shapes are generated. Figure \ref{fig:rel_vol_sq} shows the statistical plot of the discretization quality for different vertex resolutions. After around 100 vertices, the error starts to be plateaued and below $10 \%$. Therefore, we choose 100 as the number of vertices for the superquadric surface. To make the comparison relatively fair, the same number of 100 vertices are chosen to discretize the closed-form Minkowski sums boundaries in each C-slice for our HRM-based planners.

%%%%%%%%%%%%%%%%%%%%%%%%%%%%%%%%%%%%%%%%%%%%%%%%%%%%%%%%%%%%%
\subsection{Results and analysis}
An ablation study for the ``bridge C-slice'' subroutine is firstly conducted, followed by benchmark studies among the proposed HRM-based and the sampling-based planners. The benchmark results include the total time and the success rate to solve different planning problems.

%%%%%%%%%%%%%%%%%%%%%%%%%%%
\subsubsection{Ablation study for ``bridge C-slice'' subroutine} \label{sec:benchmark:result:ablation}
In this study, the HRM planner is treated as the baseline because of its deterministic property. The ablated version replaces the bridge C-slice with direct interpolation between two vertices in different C-slices and collision detection at each intermediate step using FCL. The number of steps is chosen as $N_{\rm point}$, the same with that in the bridge C-slice process. The average planning time and the number of edges in the graph (\ie $N_{\rm edge}$) are shown in Tab. \ref{tab:ablation_study}.

\begin{table}[!t]
\centering
\caption{Results of ablation study for ``bridge C-slice''}
\label{tab:ablation_study}

\begin{tabular}{cccccc}
    \hline
   Map & Robot & HRM version & Total time (s) & $N_{\rm edge}$ \\ \hline
   \multirow{2}{*}{Sparse} & \multirow{2}{*}{Rabbit} & Original & 0.2779 & 1883 \\
    & & Ablated & 0.4530 & 1959 \\
   \multirow{2}{*}{Cluttered} & \multirow{2}{*}{Rabbit} & Original & 2.238 & 10421 \\
    & & Ablated & 5.203 & 11206 \\
   \multirow{2}{*}{Maze} & \multirow{2}{*}{Rabbit} & Original & 0.8925 & 2494 \\
    & & Ablated & 1.123 & 2840 \\
   \multirow{2}{*}{Home} & \multirow{2}{*}{Chair} & Original & 87.31 & 72063 \\
    & & Ablated & 153.3 & 72785 \\
   \hline
\end{tabular}
\end{table}

The original HRM with ``bridge C-slice'' connects less valid edges than the ablated version which uses direct interpolation and explicit collision detection. This is mainly due to the fact that the computation of TFE for each robot part is conservative. However, the efficient computations of Minkowski sums for TFEs and point inclusion queries in the bridge C-slice help speed up the planner. Especially in the more complex environments like cluttered and home maps, the proposed HRM runs around two times faster than the ablated version. 

%%%%%%%%%%%%%%%%%%%%%%%%%%%
\subsubsection{Benchmark results for SE(3) and higher dimensional planning problems} \label{sec:benchmark:result:hrm}
\begin{figure*} [!t]
	\centering
	\subfloat[Run time: 3D sparse, rabbit]{
		\label{fig:bench:time:sq:sparse:rabbit}
		\centering
		\includegraphics[scale=0.48]{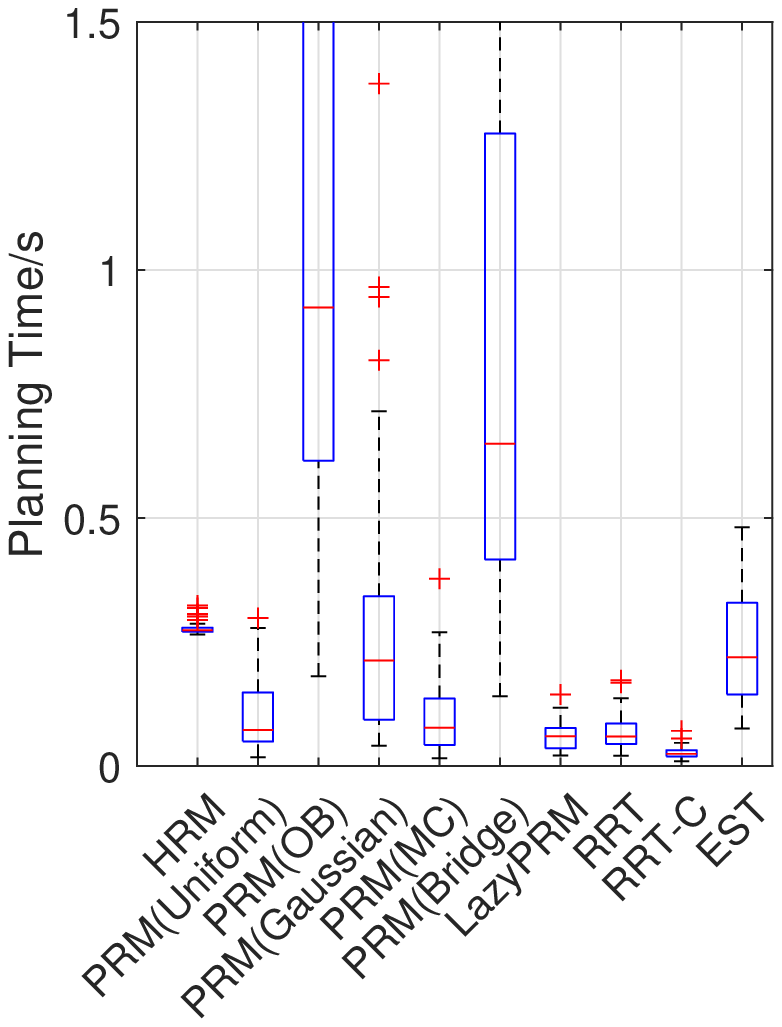}}
	\hspace{0.05in}
	\subfloat[Run time: 3D cluttered, rabbit]{
		\label{fig:bench:time:sq:cluttered:rabbit}
		\centering
		\includegraphics[scale=0.48]{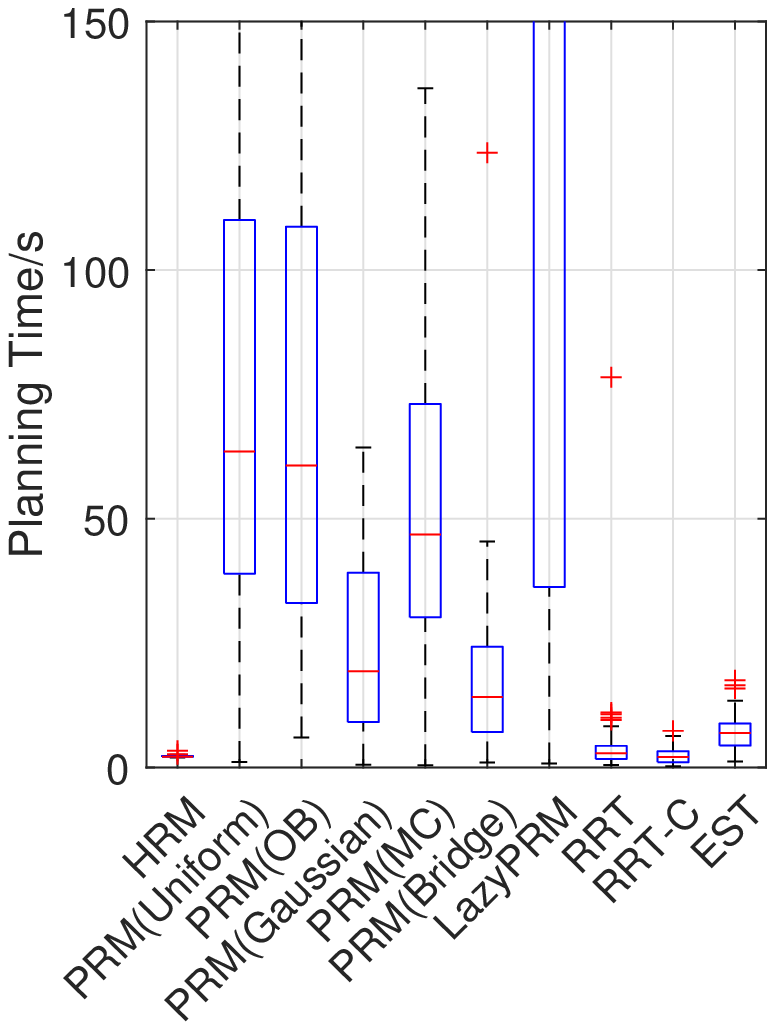}}
	\hspace{0.05in}
	\subfloat[Run time: 3D maze, rabbit]{
		\label{fig:bench:time:sq:maze:rabbit}
		\centering
		\includegraphics[scale=0.48]{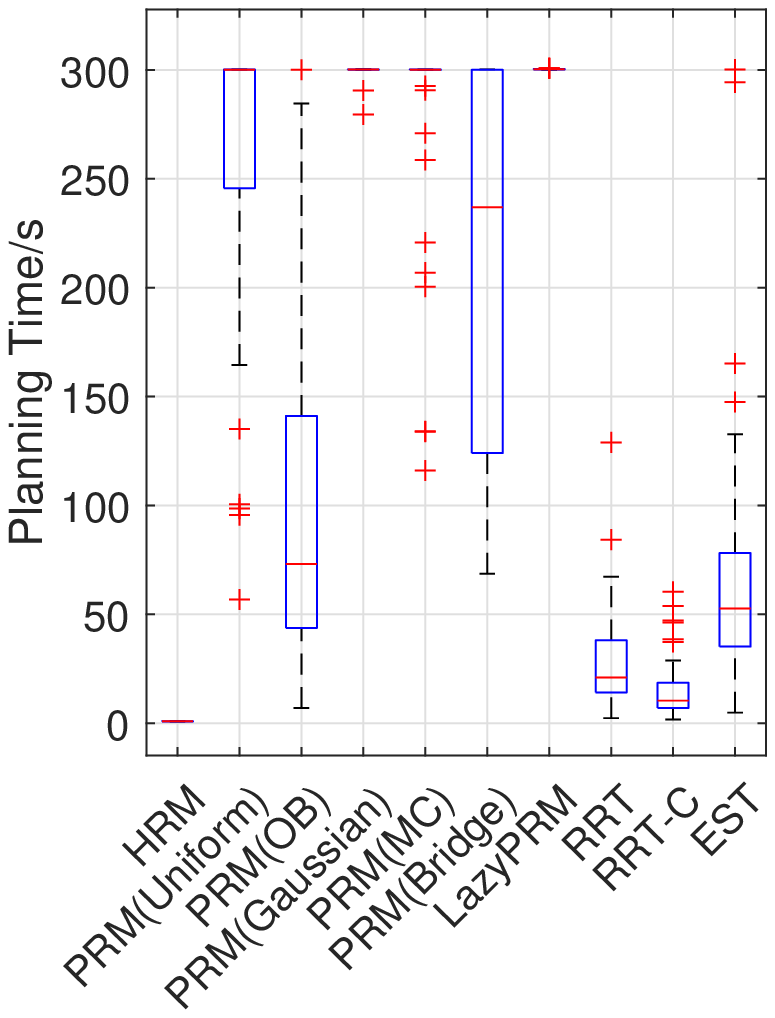}}
	\hspace{0.05in}
	\subfloat[Run time: 3D home, chair]{
		\label{fig:bench:time:sq:home:chair}
		\centering
		\includegraphics[scale=0.48]{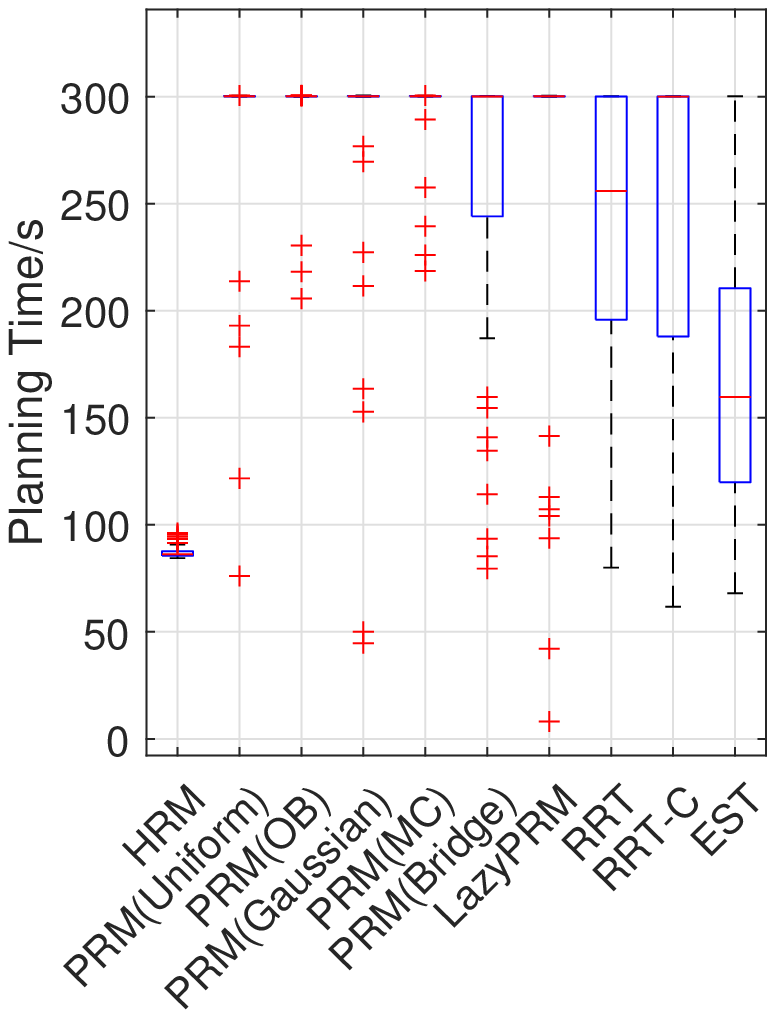}}
	\hspace{0.05in}
	\subfloat[Success rate: 3D sparse, rabbit]{
		\label{fig:bench:succ-rate:sq:sparse:rabbit}
		\centering
		\includegraphics[scale=0.48]{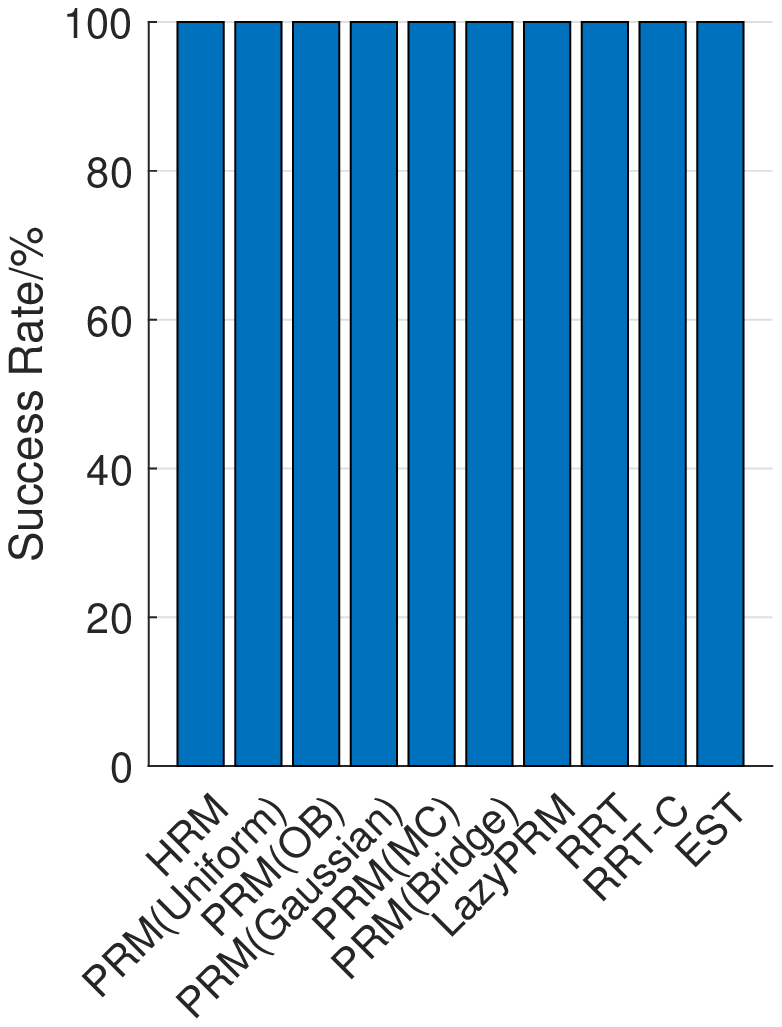}}
	\hspace{0.05in}
	\subfloat[Success rate: 3D cluttered, rabbit]{
		\label{fig:bench:succ-rate:sq:cluttered:rabbit}
		\centering
		\includegraphics[scale=0.48]{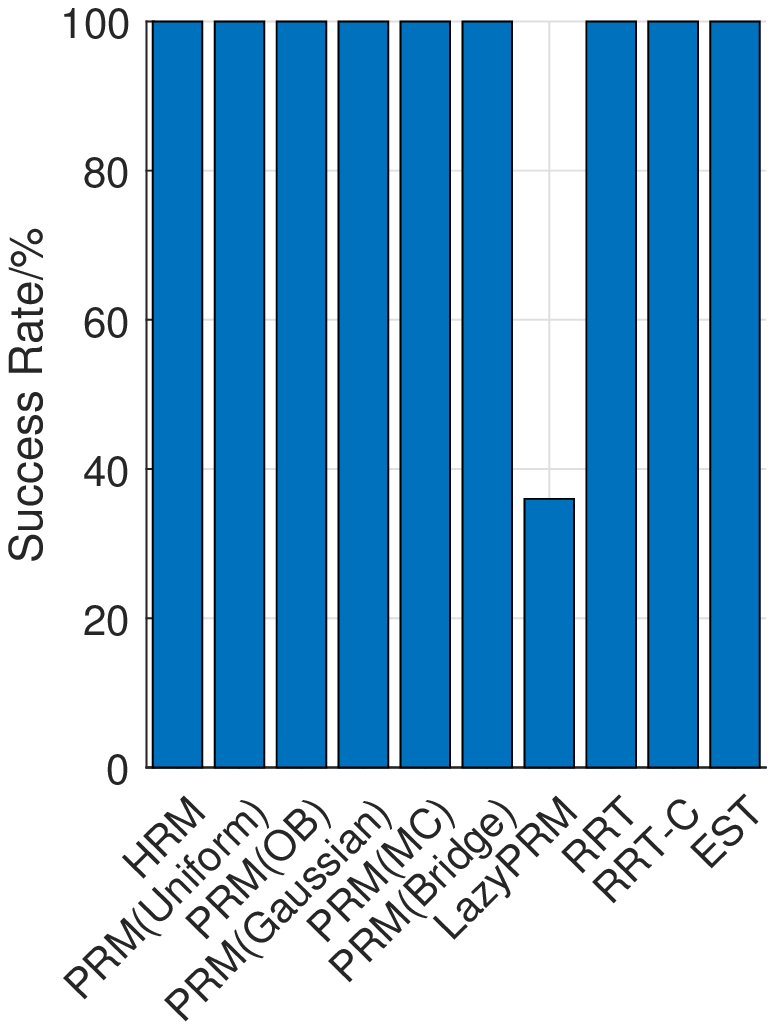}}
	\hspace{0.05in}
	\subfloat[Success rate: 3D maze, rabbit]{
		\label{fig:bench:succ-rate:sq:maze:rabbit}
		\centering
		\includegraphics[scale=0.48]{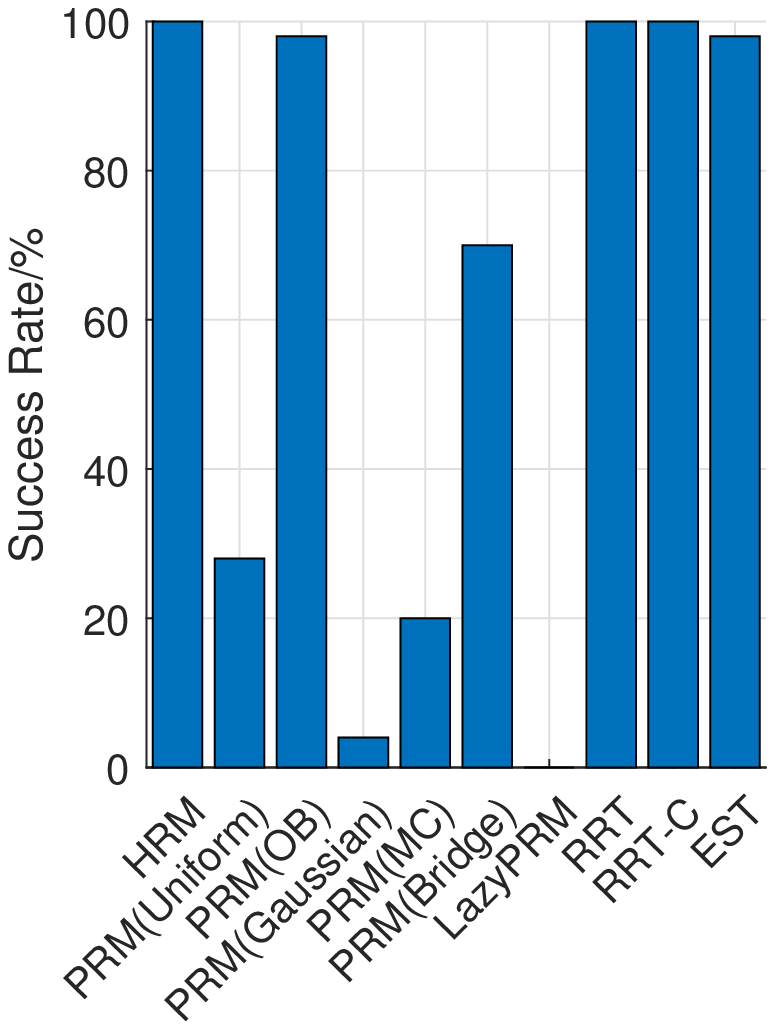}}
	\hspace{0.05in}
	\subfloat[Success rate: 3D home, chair]{
		\label{fig:bench:succ-rate:sq:home:chair}
		\centering
		\includegraphics[scale=0.48]{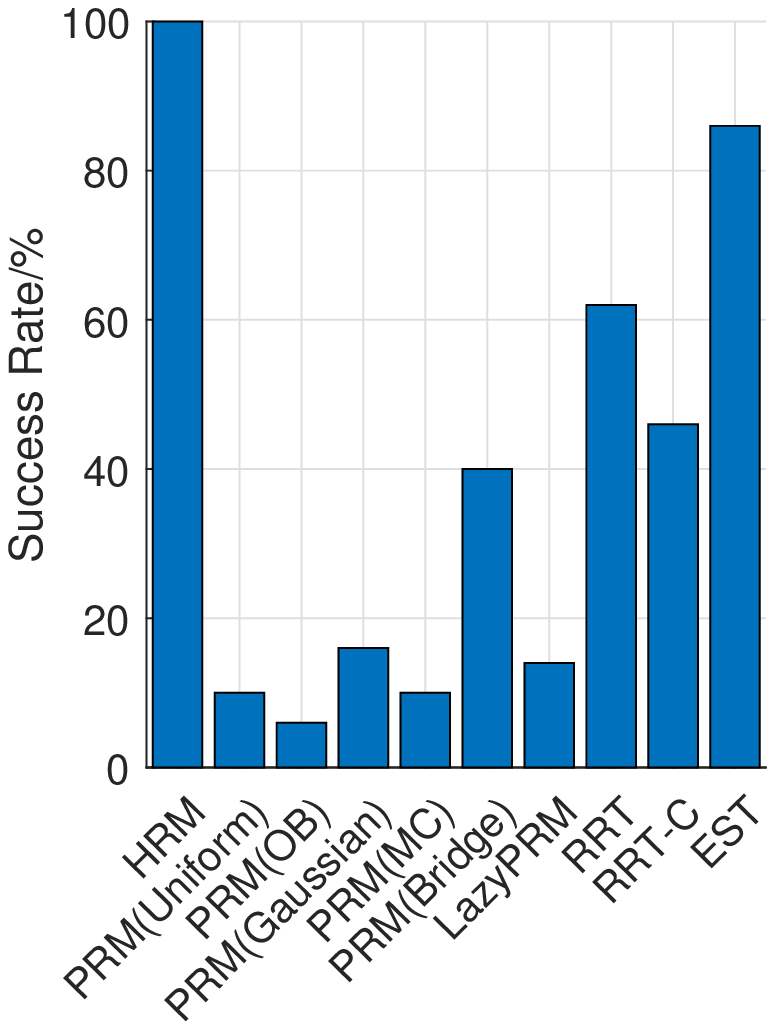}}
	\caption{Running time and success rate comparisons between HRM and sampling-based motion planners. PRM-based planners use different sampling strategies, denoted as ``PRM (sampler name)''. The planning time is shown as a box plot (Figs. \ref{fig:bench:time:sq:sparse:rabbit}, \ref{fig:bench:time:sq:cluttered:rabbit}, \ref{fig:bench:time:sq:maze:rabbit} and \ref{fig:bench:time:sq:home:chair}). The red line inside the each box is the median of the data, while the upper and lower edges of the box show the $25 \%$-th and $75 \%$-th percentile respectively. The dashed lines extend to the most extreme data points excluding the outliers. And the outliers are plotted as $+$ signs. The success rates are shown as bar plots (Figs. \ref{fig:bench:succ-rate:sq:sparse:rabbit}, \ref{fig:bench:succ-rate:sq:cluttered:rabbit}, \ref{fig:bench:succ-rate:sq:maze:rabbit} and \ref{fig:bench:succ-rate:sq:home:chair}).}
	\label{fig:bench:result:hrm}
\end{figure*}

\begin{figure*} [!t]
	\centering
	\subfloat[3D sparse, snake]{
		\label{fig:bench:time:sq:sparse:snake}
		\centering
		\includegraphics[scale=0.48]{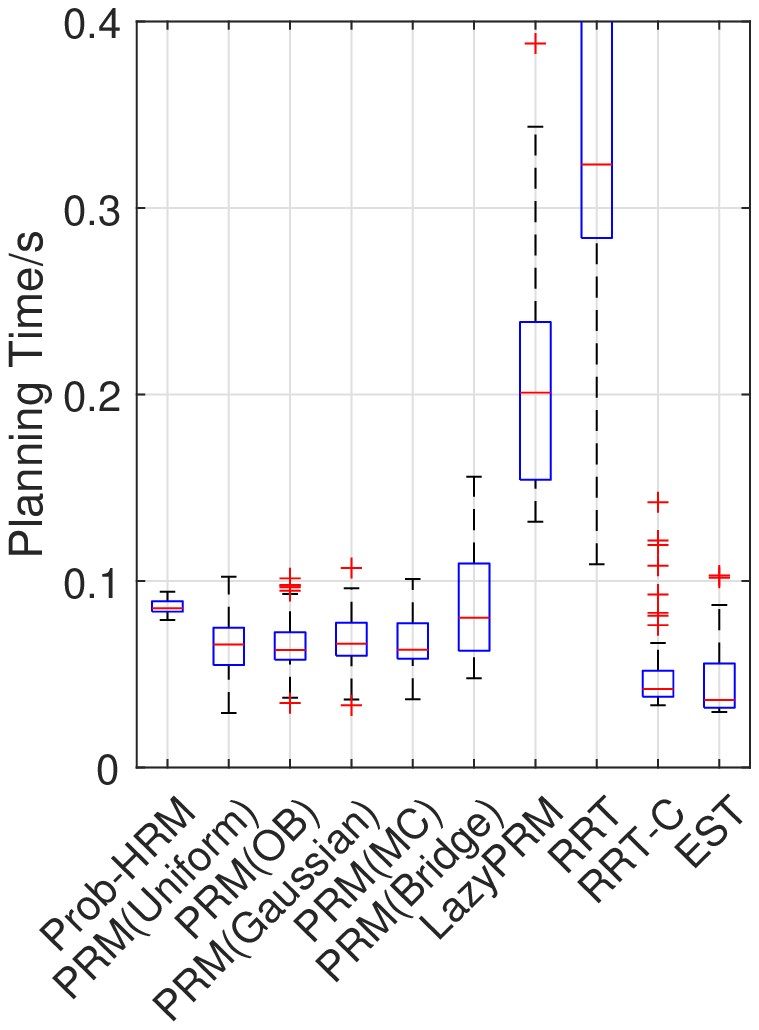}}
	\hspace{0.05in}
	\subfloat[3D sparse, tree]{
		\label{fig:bench:time:sq:sparse:tree}
		\centering
		\includegraphics[scale=0.48]{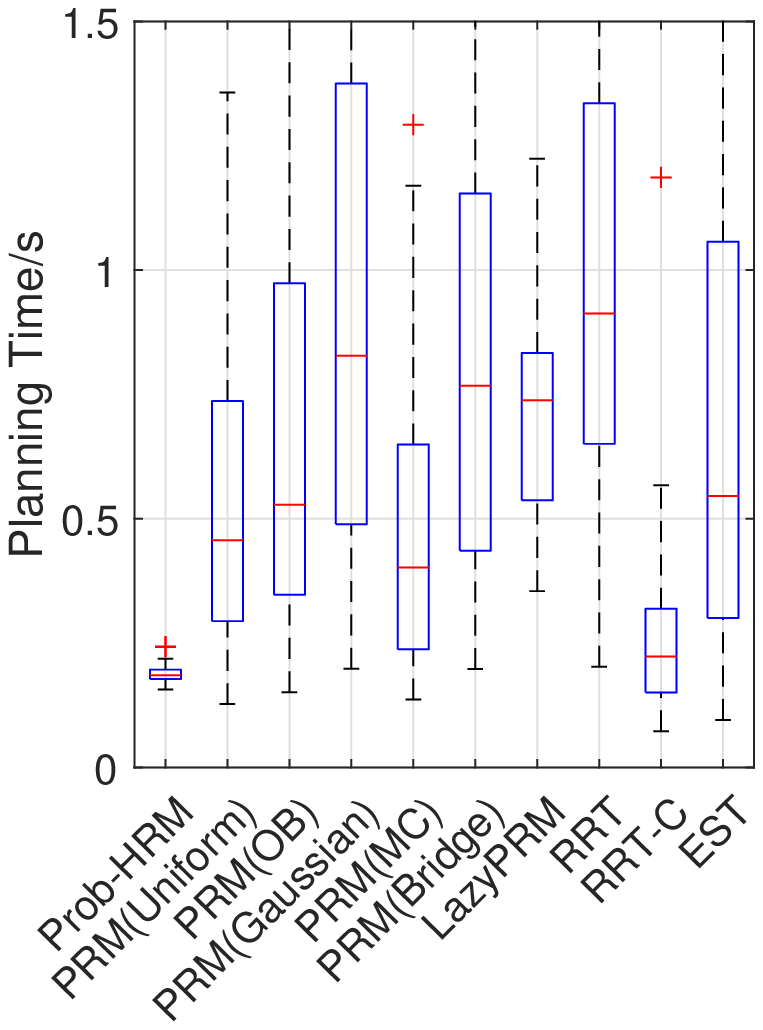}}
	\hspace{0.05in}
	\subfloat[3D cluttered, snake]{
		\label{fig:bench:time:sq:cluttered:snake}
		\centering
		\includegraphics[scale=0.48]{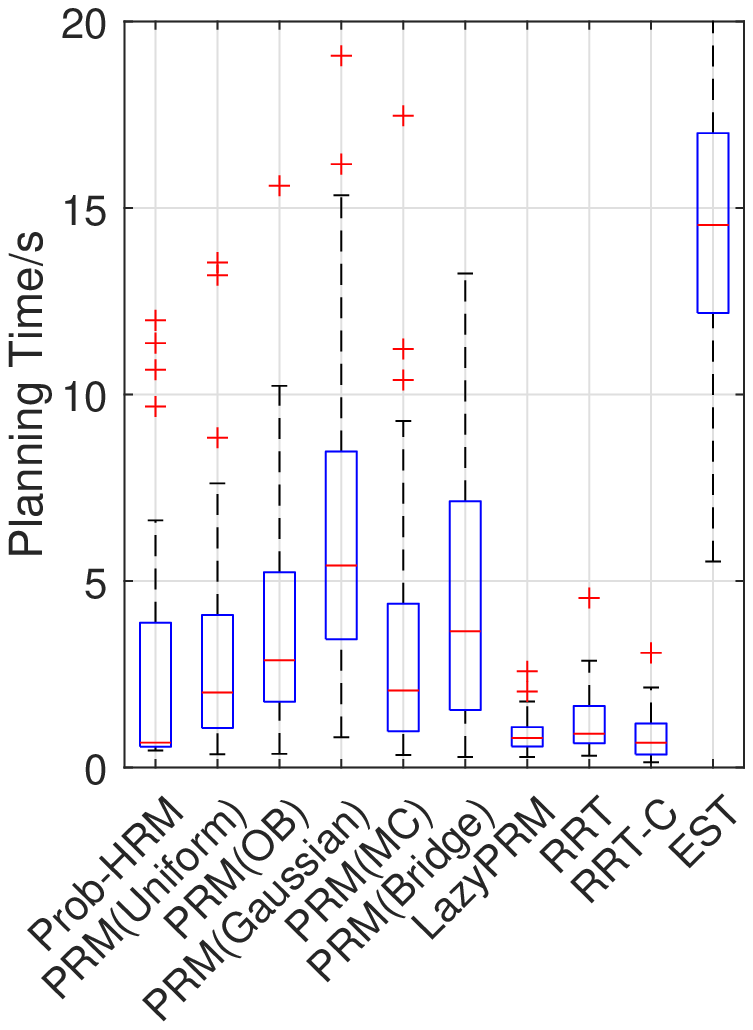}}
	\hspace{0.05in}
	\subfloat[3D cluttered, tree]{
		\label{fig:bench:time:sq:cluttered:tree}
		\centering
		\includegraphics[scale=0.48]{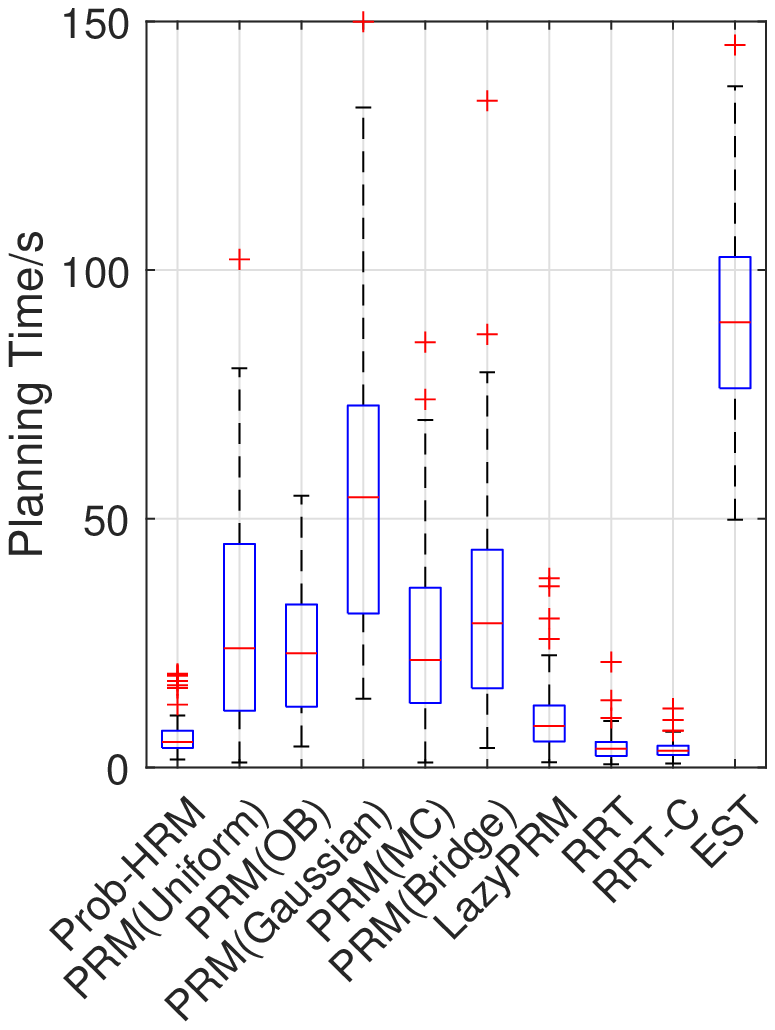}}
	\hspace{0.05in}
	\subfloat[3D maze, snake]{
		\label{fig:bench:time:sq:maze:snake}
		\centering
		\includegraphics[scale=0.48]{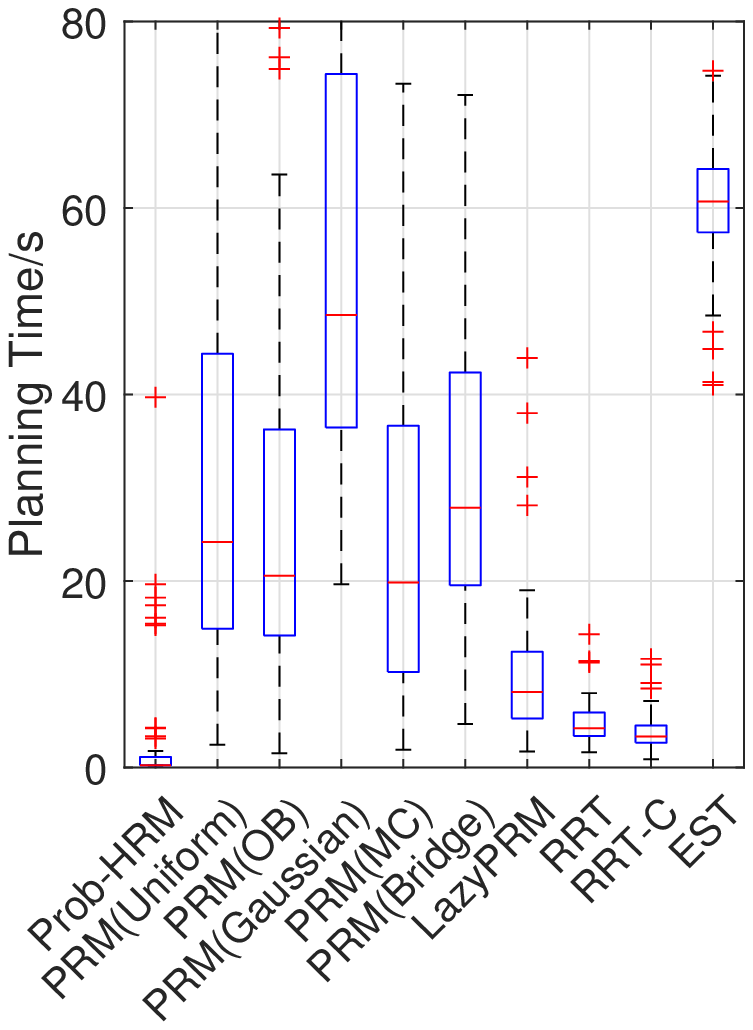}}
	\hspace{0.05in}
	\subfloat[3D home, snake]{
		\label{fig:bench:time:sq:home:snake}
		\centering
		\includegraphics[scale=0.48]{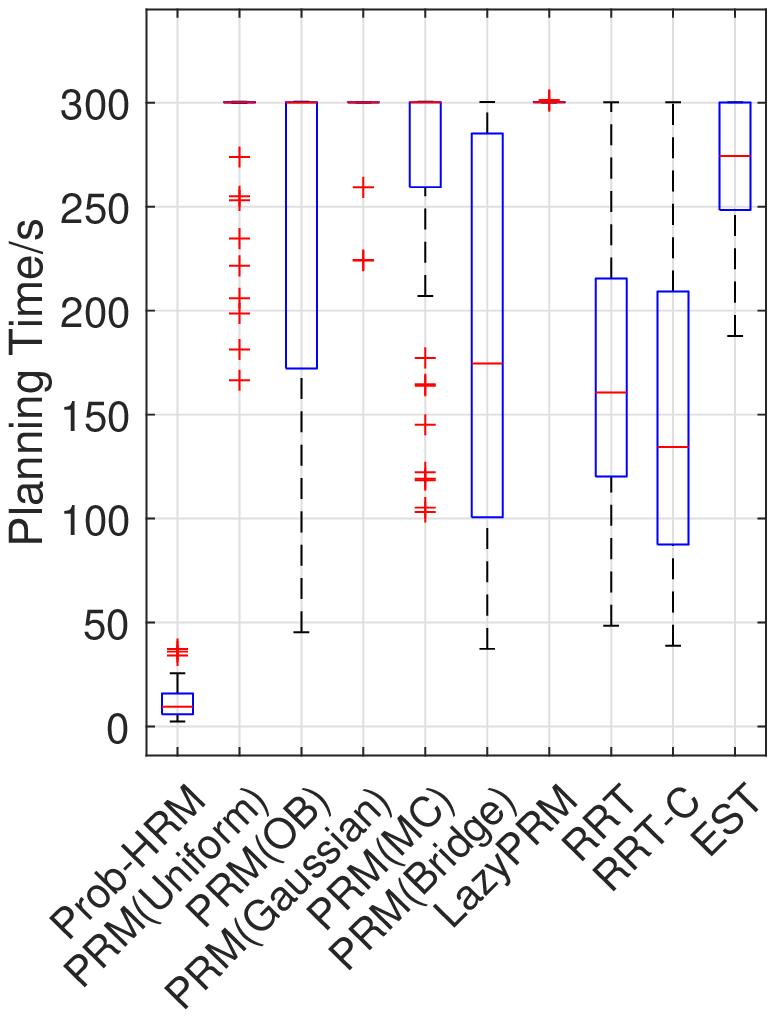}}
	\hspace{0.05in}
	\subfloat[3D narrow, snake]{
		\label{fig:bench:time:sq:narrow:snake}
		\centering
		\includegraphics[scale=0.48]{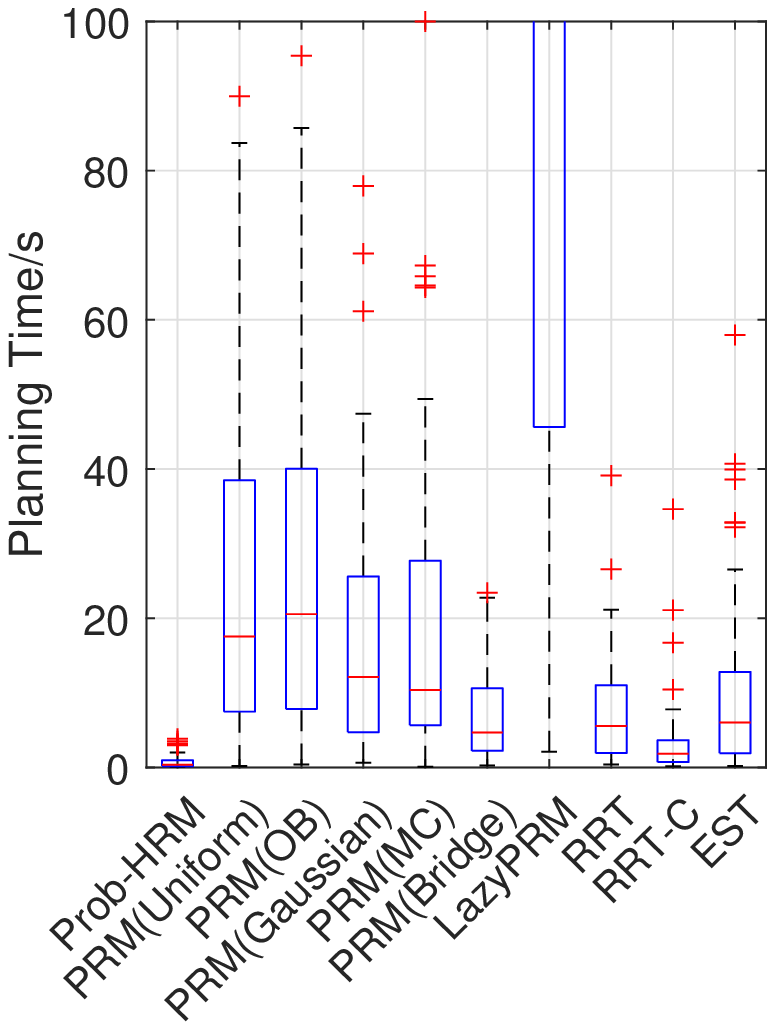}}
	\hspace{0.05in}
	\subfloat[3D narrow, tree]{
		\label{fig:bench:time:sq:narrow:tree}
		\centering
		\includegraphics[scale=0.48]{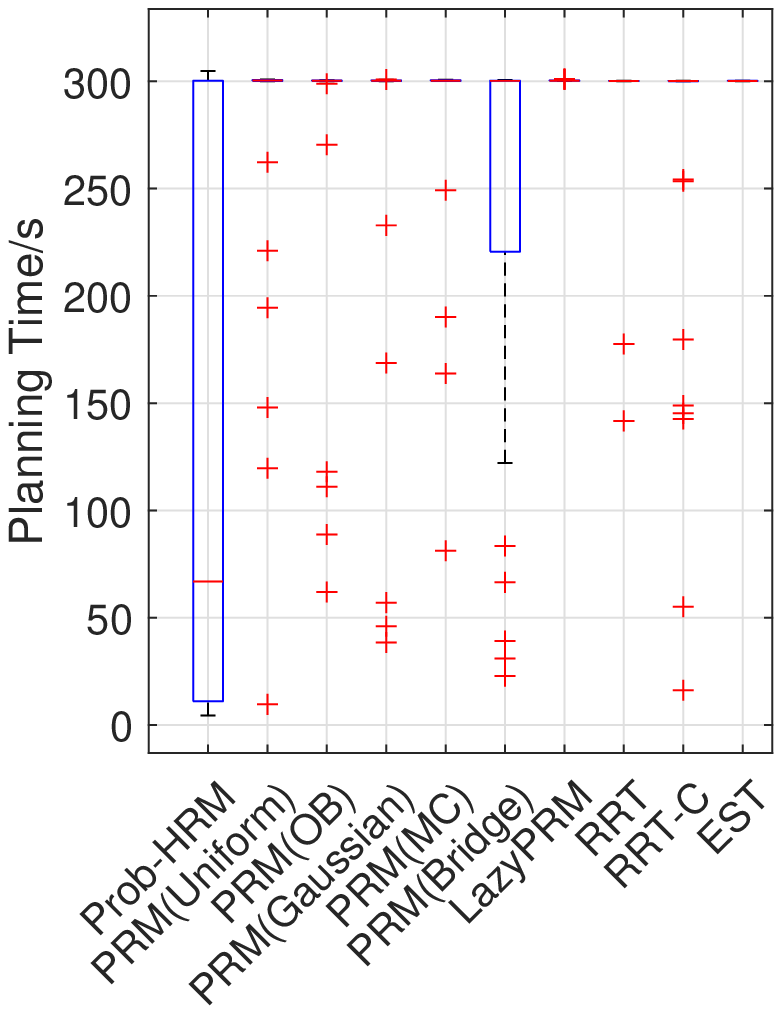}}
		
	\caption{Running time comparisons between Prob-HRM and sampling-based motion planners.}
	\label{fig:bench:result:prob-hrm:time}
\end{figure*}

\begin{figure*} [!t]
	\centering
	\subfloat[3D sparse, snake]{
		\label{fig:bench:succ-rate:sq:sparse:snake}
		\centering
		\includegraphics[scale=0.48]{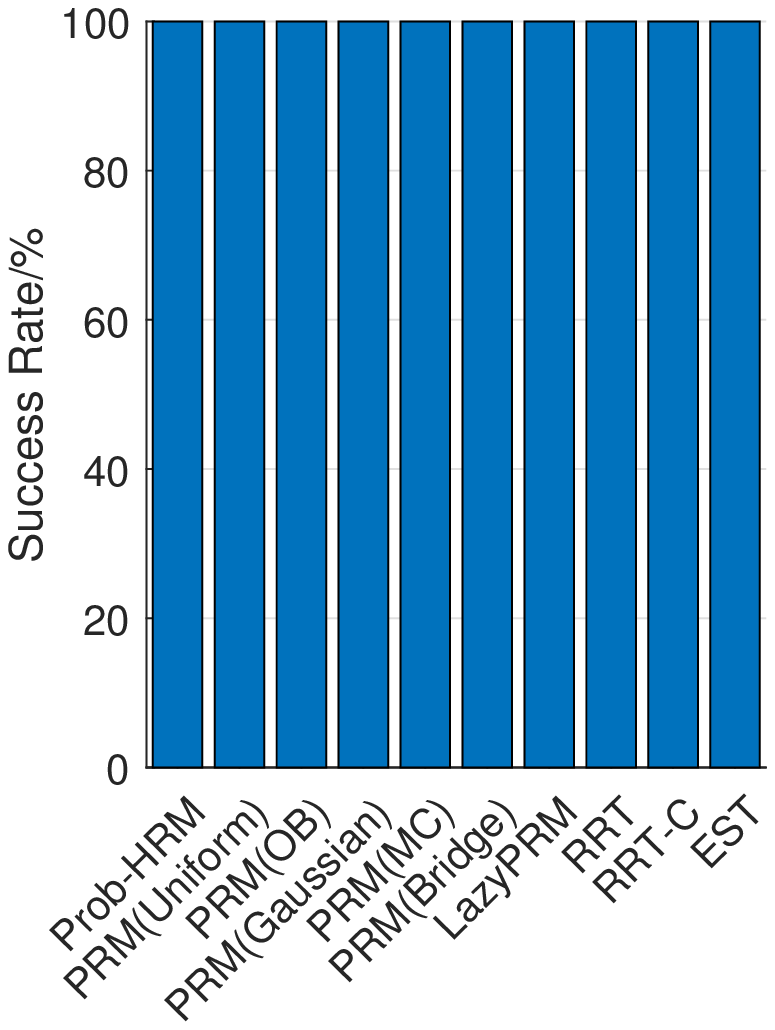}}
	\hspace{0.05in}
	\subfloat[3D sparse, tree]{
		\label{fig:bench:succ-rate:sq:sparse:tree}
		\centering
		\includegraphics[scale=0.48]{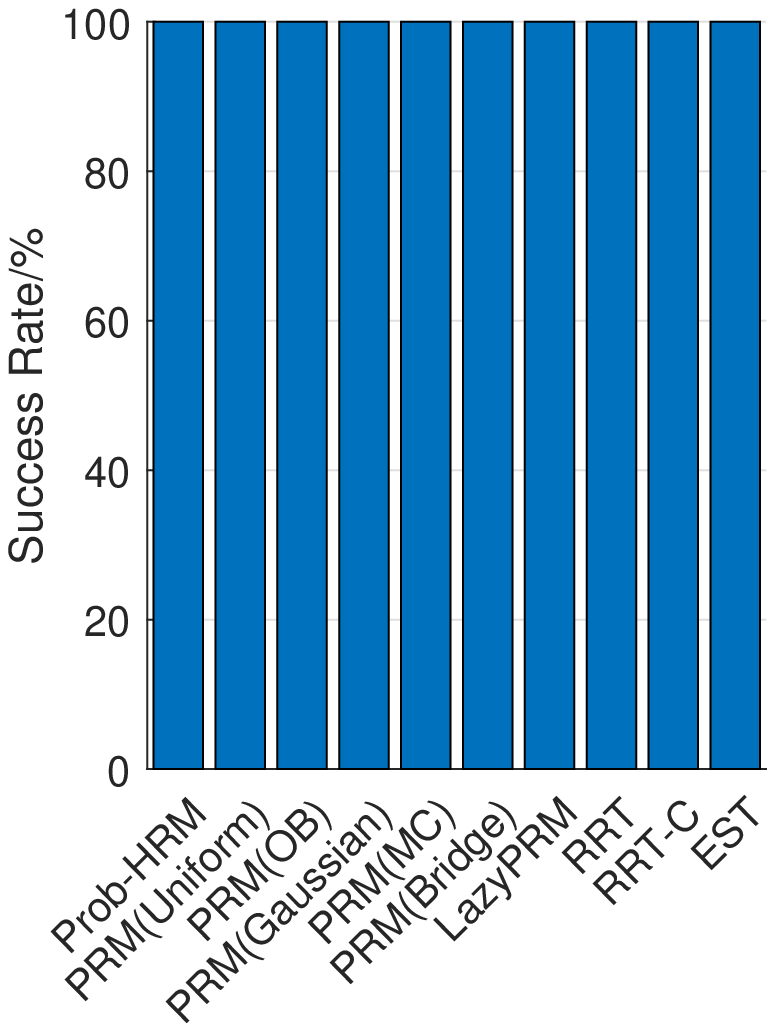}}
	\hspace{0.05in}
	\subfloat[3D cluttered, snake]{
		\label{fig:bench:succ-rate:sq:cluttered:snake}
		\centering
		\includegraphics[scale=0.48]{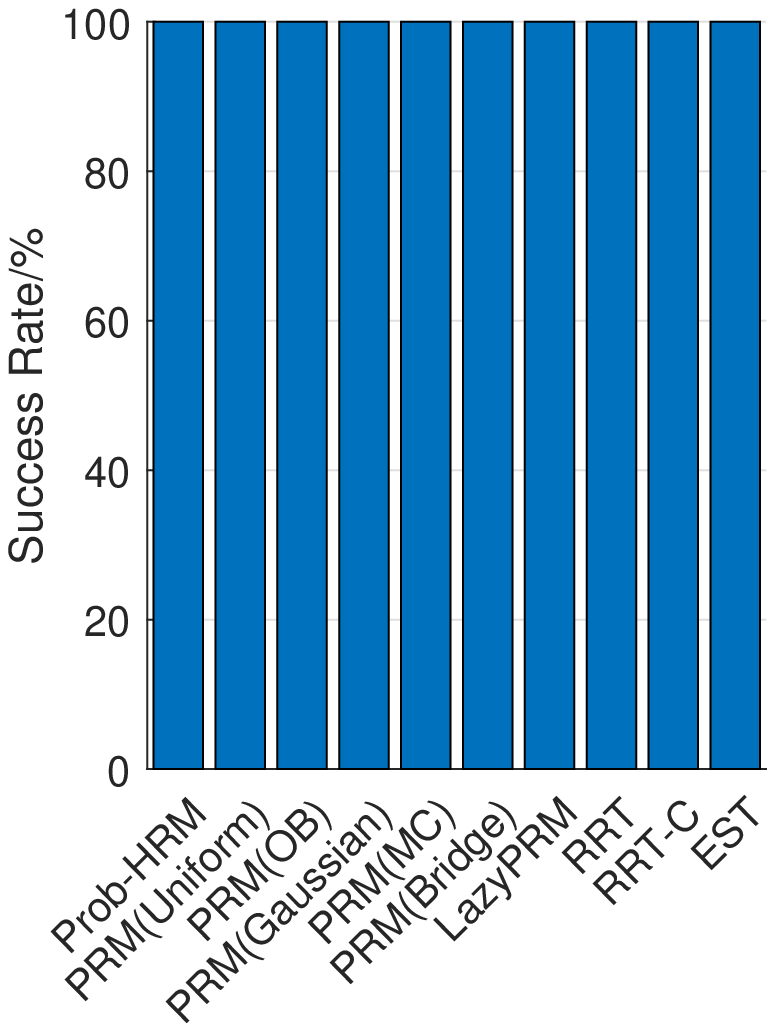}}
	\hspace{0.05in}
	\subfloat[3D cluttered, tree]{
		\label{fig:bench:succ-rate:sq:cluttered:tree}
		\centering
		\includegraphics[scale=0.48]{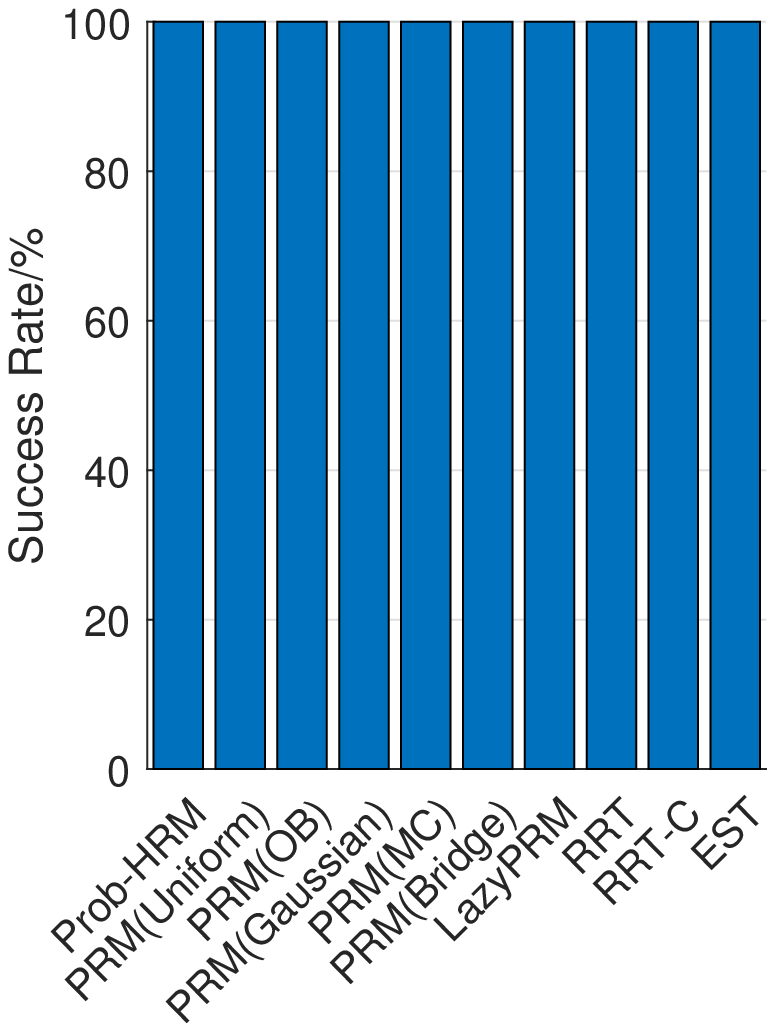}}
	\hspace{0.05in}
	\subfloat[3D maze, snake]{
		\label{fig:bench:succ-rate:sq:maze:snake}
		\centering
		\includegraphics[scale=0.48]{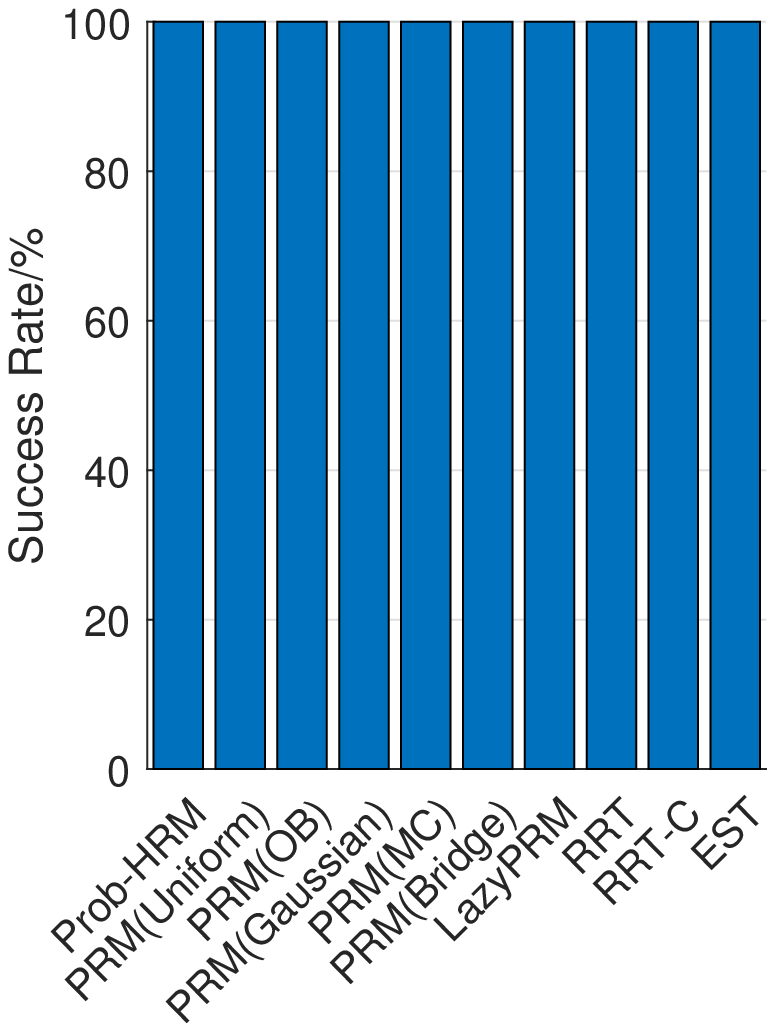}}
	\hspace{0.05in}
	\subfloat[3D home, snake]{
		\label{fig:bench:succ-rate:sq:home:snake}
		\centering
		\includegraphics[scale=0.48]{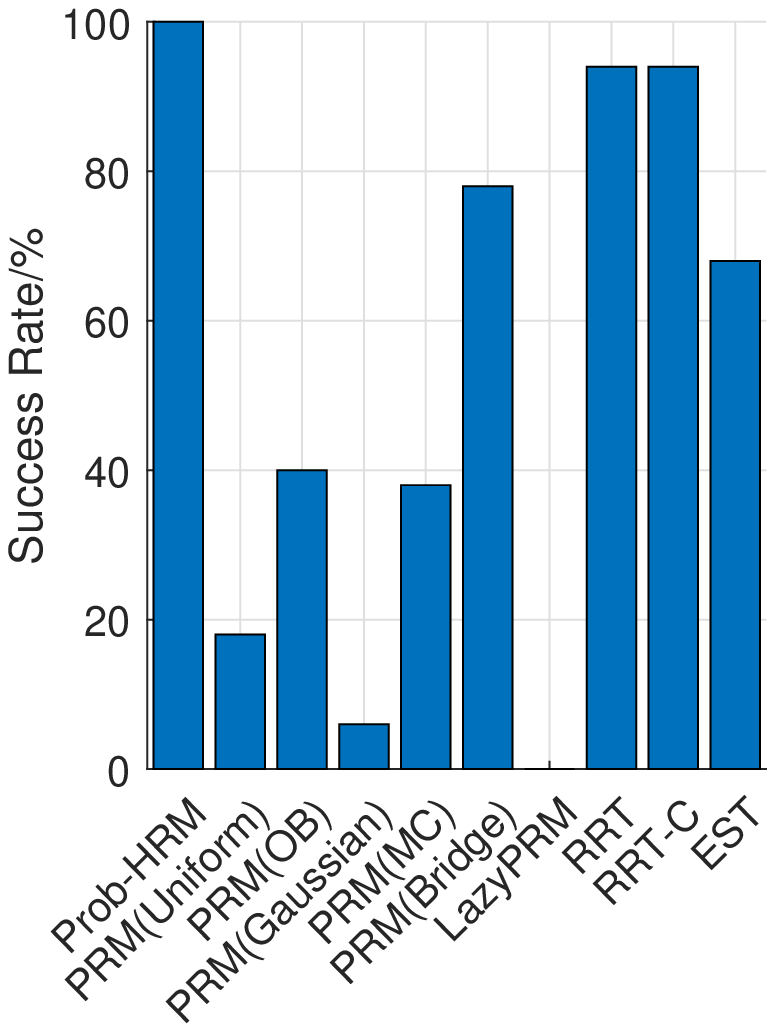}}
	\hspace{0.05in}
	\subfloat[3D narrow, snake]{
		\label{fig:bench:succ-rate:sq:narrow:snake}
		\centering
		\includegraphics[scale=0.48]{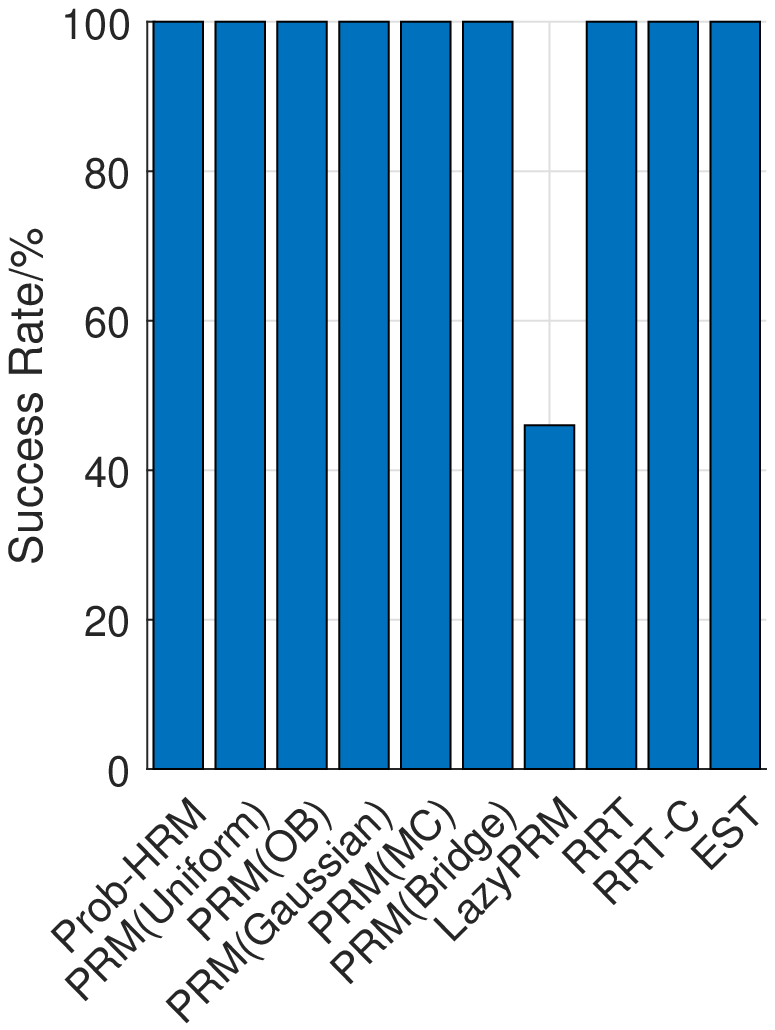}}
	\hspace{0.05in}
	\subfloat[3D narrow, tree]{
		\label{fig:bench:succ-rate:sq:narrow:tree}
		\centering
		\includegraphics[scale=0.48]{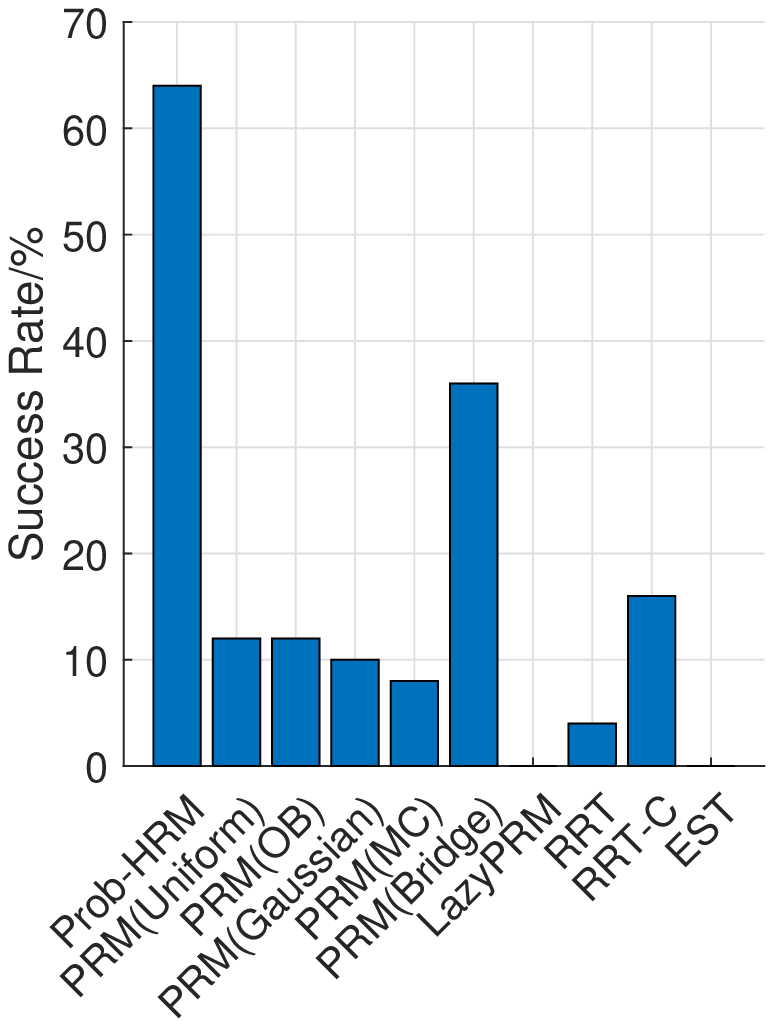}}
		
	\caption{Success rate comparisons between Prob-HRM and sampled-based motion planners.}
	\label{fig:bench:result:prob-hrm:sr}
\end{figure*}

The comparisons of total running time and success rate in $\SE(3)$ rigid-body planning problems are shown in Fig. \ref{fig:bench:result:hrm}. Figures \ref{fig:bench:result:prob-hrm:time} and \ref{fig:bench:result:prob-hrm:sr} show the computational time and success rate results for articulated robots in $\SE(3) \times (S^1)^n$ configuration space, respectively. For our proposed HRM-based and the PRM-based planners, the total running time at each trial includes both graph construction and search phases. 

% Table \ref{tab:bench:result:hrm:datastructure} shows the comparisons of the graph/tree sizes between HRM and sampling-based planners in two typical scenes, \ie rabbit robot in sparse map and chair object in home map. The resulting numbers of vertices and edges information are averaged and rounded to the nearest integer. 

% \begin{table} [!t]
% 	\centering
% 	\caption{Comparisons of the size of data structures for Prob-HRM and sampling-based planners}
% 	\label{tab:bench:result:prob-hrm:datastructure}
	
% 	\begin{tabular}{cc|cccc}
% 		\hline
% 		Map (Robot) & Planner & $N_{\rm vertex}$ & $N_{\rm edge}$ & $N_{\rm path}$ \\ \hline
% 		Cluttered (Snake) & Prob-HRM & 7871 & 14159 & 49 \\
% 		Cluttered (Snake) & PRM & 76 & 398 & 9 \\
% 		Cluttered (Snake) & Lazy PRM & 257 & 1992 & 14 \\
% 		Cluttered (Snake) & RRT & 53 & 52 & 10 \\
% 		Cluttered (Snake) & RRT Connect & 34 & 33 & 10 \\
% 		Cluttered (Snake) & EST & 487 & 486 & 21 \\ 
% 		\hline
% 		Narrow (Tree) & Prob-HRM & 46931 & 111150 & 21 \\
% 		Narrow (Tree) & PRM & 1725 & 11698 & 16 \\
% 		Narrow (Tree) & Lazy PRM & 31055 & 298000 & -- \\
% 		Narrow (Tree) & RRT & 4104 & 4103 & 10 \\
% 		Narrow (Tree) & RRT Connect & 3760 & 3759 & 14 \\
% 		Narrow (Tree) & EST & 2822 & 2821 & -- \\
% 		\hline
% 	\end{tabular}
% \end{table}

From the benchmark results, sampling-based planners are very efficient when the environments are sparse (such as in Figs. \ref{fig:bench:time:sq:sparse:rabbit}, \ref{fig:bench:time:sq:sparse:snake}, \ref{fig:bench:time:sq:sparse:tree}, etc). However, they become slower when the space occupied by obstacles increases. Also, the success rate of sampling-based planners decreases as the environment becomes denser. In cases like in Figs. \ref{fig:bench:succ-rate:sq:home:snake} and \ref{fig:bench:succ-rate:sq:narrow:tree}, some planners cannot even find any solution within the assigned time limit of 300 seconds. For graph-based algorithms, even with the help of different types of samplers, they still take longer time to finally find a valid path. The tree-based planners are much more efficient for single queries in sparse and cluttered maps. And even in the maze map, when the dimensions of the problems increase, both RRT and RRT-connect planners can still search for a valid path efficiently. But in more complex maps like the home and narrow environments, both of their speed and success rate start to drop.

On the other hand, the proposed HRM and Prob-HRM planners are more efficient in complex environments, such as in Figs. \ref{fig:bench:time:sq:maze:rabbit}, \ref{fig:bench:time:sq:home:chair}, \ref{fig:bench:time:sq:home:snake} and \ref{fig:bench:time:sq:narrow:tree}. The success rates among multiple planning trials are also higher, as in Figs. \ref{fig:bench:succ-rate:sq:maze:rabbit}, \ref{fig:bench:succ-rate:sq:home:chair}, \ref{fig:bench:succ-rate:sq:home:snake} and \ref{fig:bench:succ-rate:sq:narrow:tree}. These results show the advantages of the proposed HRM-based planners in solving narrow passage problems. Furthermore, as can be seen from Fig. \ref{fig:bench:result:hrm}, the performance of HRM is more stable among different trials in rigid-body planning problems, which is mainly due to its deterministic nature. Prob-HRM planner, on the other hand, has larger variance in planning time for articulated robots (such as in Fig. \ref{fig:bench:time:sq:cluttered:snake}). Another feature of our proposed HRM and Prob-HRM is that they are both graph-based planners. They are competitive in solving complex problems with single-query planners (as in Figs. \ref{fig:bench:time:sq:cluttered:rabbit}, \ref{fig:bench:time:sq:maze:snake} and \ref{fig:bench:time:sq:narrow:snake}), and outperforms all planners in environments with narrow corridors (as in Figs \ref{fig:bench:time:sq:home:chair}, \ref{fig:bench:time:sq:home:snake} and \ref{fig:bench:time:sq:narrow:tree}). This is desirable since ours can not only build the roadmap efficiently but also answer planning queries multiple times when the environment does not change. 

% Prob-HRM can also solve higher dimensional problems via randomly sampling the robot shapes.

% Physical Experiments
\section{Physical Experiments on Walking Path Planning for a Humanoid Robot}
\label{sec:real-experiment}
\begin{figure*} [!t]
    \centering
    \includegraphics[scale=0.43]{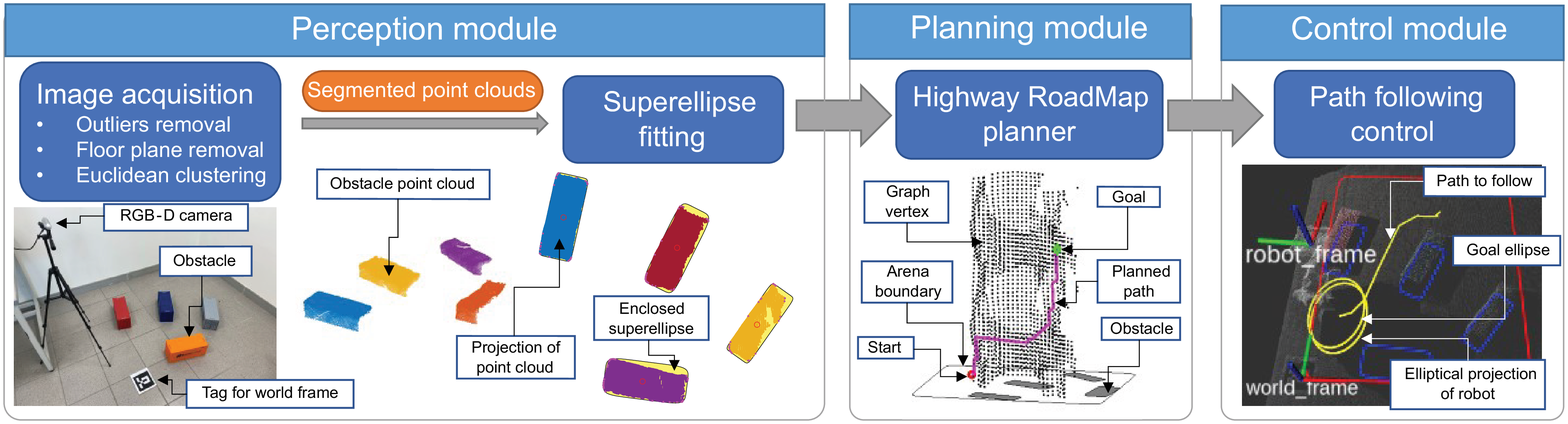}
    \caption{The pipeline for physical experiments of the walking path planning for NAO humanoid robot.}
    
    \label{fig:exp:pipeline}
\end{figure*}

In order to demonstrate the capabilities of our proposed planning framework in the real-world setting, physical experiments with a NAO humanoid robot \cite{nao2021website} are conducted. The task is to guide the robot to walk through environments with several objects on the floor in random poses. The robot is required to avoid them in order to pass this cluttered space. Therefore, the problem is simplified into a planar case, where the robot and all objects are projected onto the floor. The contour of the robot projection is encapsulated by an ellipse, with pre-defined semi-axes lengths. The robot is able to walk sideways and its the configuration space is $\SE(2)$. The arena is a pre-defined rectangular area, which is bounded by a superellipse with exponent defined as $0.1$. The whole experimental pipeline consists of three main modules: perception, planning, and control, which is shown in Fig \ref{fig:exp:pipeline}. Robot Operating System (ROS) is used to communicate between different modules.

The whole scene is firstly captured from a fixed RGB-D camera as point cloud data. The point cloud is transformed from the camera frame into the world frame (indicated by an ArUco marker \cite{garrido2014automatic} on the floor), and segmented into disjoint clusters using Point Cloud Library (PCL) \cite{rusu20113d}. Each cluster is then projected onto the x-y plane and fitted into a superelliptical model using Eq. \eqref{eq:sq_fitting_2d}. The obtained environmental data is then given as the input to the planning module. By manually selecting the start and goal poses of the robot, a valid $\SE(2)$ path is then solved by the proposed HRM planner. Finally, given a list of $\SE(2)$ poses, the robot follows the path via a simple proportional controller \cite{han2020can}. The robot pose is tracked by an ArUco tag attached to its head and is controlled to minimize the distance with the next way point on the trajectory until reaching the goal configuration.

\begin{figure*}[!t]
    \centering
    \includegraphics[scale = 0.18]{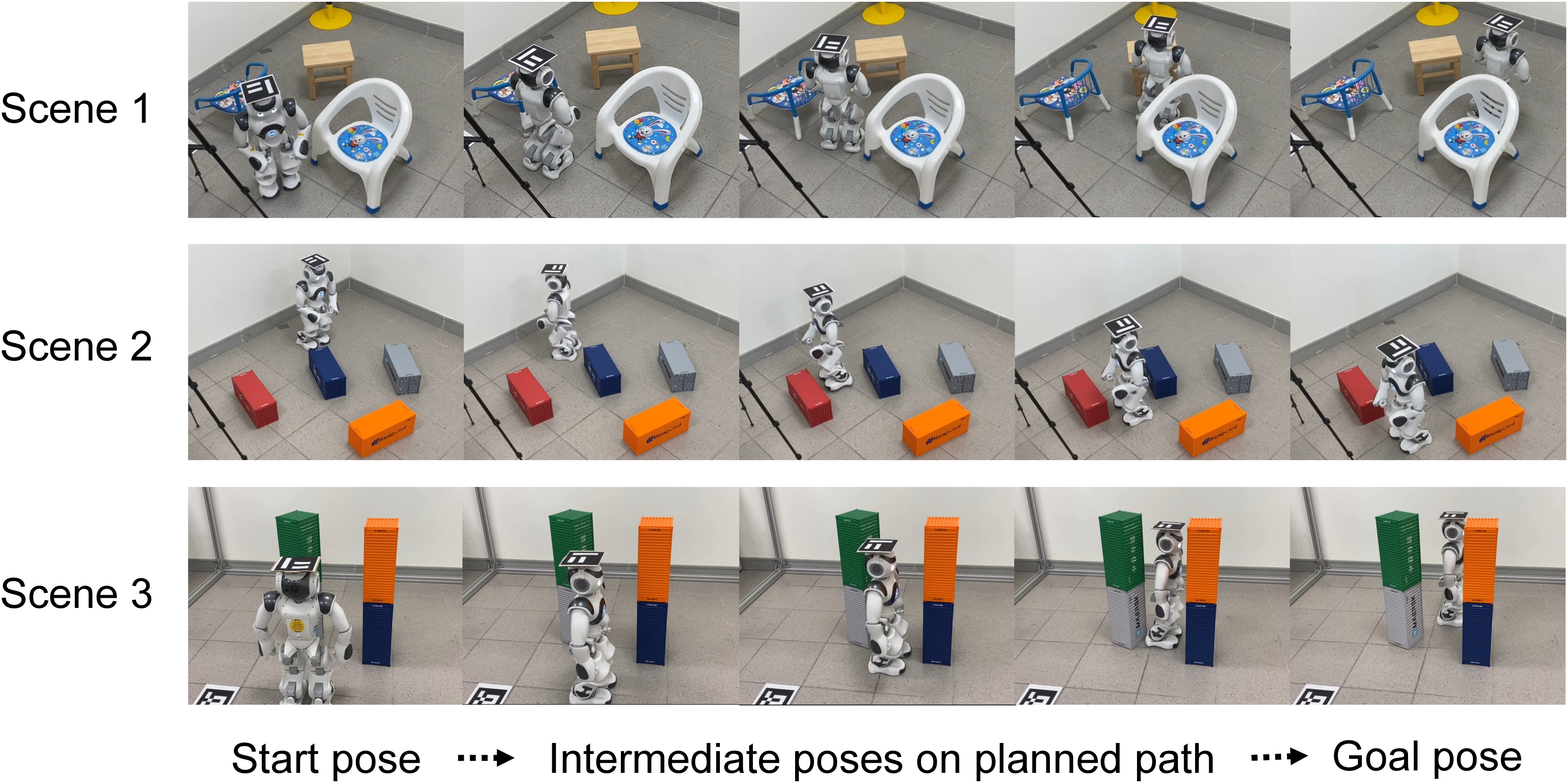}
    \caption{Walking sequences of NAO that follows the planned $\SE(2)$ paths in physical experiments.}
    \label{fig:exp:screenshot}
\end{figure*}

Since the planning scene does not change during the whole trial of the experiment, the perception and planning modules both run offline. The control module runs as an on-board process to keep the robot following the solved path. Table \ref{tab:exp:results} shows the planning results in different example trials of experiments. Figure \ref{fig:exp:screenshot} demonstrates the walking sequences of NAO for the three different planning scenarios.

\begin{table}[!t]
\centering
\caption{NAO walking path planning results using HRM}
\label{tab:exp:results}

\begin{tabular}{cccc}
\hline
    Scene & Graph time (ms) & Search time (ms) & Total time (ms) \\ \hline
    1 & 55.71 & 3.18 & 58.90 \\
    2 & 17.62 & 1.07 & 18.69 \\
    3 & 147.46 & 4.53 & 151.99 \\
    \hline
\end{tabular}
\end{table}

% Discussion
\section{Discussions}
\label{sec:discussion}
This section discusses the advantages of the proposed HRM-based planners, followed by some potential limitations.

%%%%%%%%%%%%%%%%%%%%%%%%%%%%%%%%%%%%%%%%%%%%%%%%%%%%%%%%%%%%%%%%%%%%%%%%%%%%%%%%%%%
\subsection{Geometric approximations of rigid objects} \label{sec:discussion:geom-approx}
The superquadric is an example model to enclose the environmental features. Alternatively, the convex polyhedron is a well-known type of geometry to represent a complex body, but may require many vertices and faces to describe a rounded region. It is possible to fit a convex polyhedra with superquadric model and vice versa, which introduces approximation errors. The fitting quality is evaluated as the relative volume between the two different models as in Eq. \eqref{eq:rel_vol_sq}.

To fit a superquadric model, the vertices of a convex polyhedron is used in Eq. \eqref{eq:sq_fitting_3d}. The evaluation metrics include not only Eq. \eqref{eq:rel_vol_sq}, but also the averaged sum of absolute difference between the point and implicit function, \ie 
$
\kappa_{\rm implicit} = \frac{1}{m} \sum_{i=1}^{m} \left| \Phi({\bf x}_i) - 1 \right|,
$
where $m$ is the number of vertices of the convex polyhedron. 

Firstly, the convex polyhedron is treated as ground truth and generated as the convex hull of a set of 100 random points. Two types of convex polyhedra are studied: centrally symmetric and random shapes. To generate the centrally symmetric convex polyhedra, all the random vertices are flipped around the origin, followed by computing the convex hull. Among all the 100 trials, the mean of $\kappa_{\rm volume}$ and $\kappa_{\rm implicit}$ are: for the centrally symmetric polyhedra, $11.74 \%$ and $0.3886$ respectively; and, for the random polyhedra, $19.88 \%$ and $0.6057$ respectively. The results show that the superquadric surfaces fit closely to the polyhedral vertices when the object is centrally symmetric. However, when the convex polyhedron is highly non-symmetric, fitting the central-symmetric superquadric model is conservative and the volume difference might be unavoidably large. On the other hand, a superquadric body can be considered as the ground truth, which this article mainly addresses and uses for benchmarks with sampling-based planners. The fitting process is introduced when selecting parameters for sampling-based planners in Sec. \ref{sec:benchmark:parameter:sampling-based}. It can be seen that a good convex polyhedral approximation uses many more sampled points. This is mainly because that a better faceted representation of the curved surface of a superquadric requires denser set of sampled points.

%%%%%%%%%%%%%%%%%%%%%%%%%%%%%%%%%%%%%%%%%%%%%%%%%%%%%%%%%%%%%%%%%%%%%%%%%%%%%%%%%%%
\subsection{Parameters selection for HRM and Prob-HRM}
Two major parameters that affect the success and performance of the proposed algorithms are the number of C-slices ($N_{\rm slice}$) and sweep lines ($N_{\rm line}$). 

For the HRM planner, a pre-defined deterministic sampling of the orientation is required, which is discussed in Sec. \ref{sec:hrm-planner:orientations}. For Prob-HRM planner, $N_{\rm slice}$ is incremental during the planning, meaning that the user does not need to provide this parameter beforehand. When there is no path found, this number will keep increasing until the termination condition is satisfied. Therefore, in this case, the roadmap can be kept refining instead of being computed from scratch again. Also, one could store the roadmap after one trial of planning for further reuse and refinements.

For both HRM and Prob-HRM planners, the selection of $N_{\rm line}$ can be either user-defined or computed based on Eq. \eqref{eq:bench:num_line}. The latter choice is the default of the proposed planners. This choice initially generates a coarse resolution of sweep lines that can efficiently solve an easy problem, but tries to detect most of the C-obstacles. Since there is a refinement step for the existing roadmap, the input $N_{\rm line}$ only defines an initial resolution of the roadmap. When the problem becomes more complex, \eg including narrow passages, the existing roadmap will be made denser by iteratively doubling the number of sweep lines until one of the termination conditions is satisfied.

%%%%%%%%%%%%%%%%%%%%%%%%%%%%%%%%%%%%%%%%%%%%%%%%%%%%%%%%%%%%%%%%%%%%%%%%%%%%%%%%%%%
\subsection{Advantageous properties of our proposed framework}
One of the highlights of our proposed path planning framework is the closed-form parameterization of Minkowski sum and difference that explicitly characterizes the C-space. The closed-form expression only depends on the parameters of one body (such as the superquadric obstacle body when computing the C-obstacle boundary). Therefore, the computational complexity is linear with respect to only one body, not both ones as the traditional polytope-based Minkowski sum \cite{fogel2009exactcomplexity}. Moreover, the numerical errors introduced in this process only come from the geometric approximations of object since the Minkowski sums computations are exact. The density of sampling vertices on the object surface is determined and used throughout all the experiments in this article. It is shown to be robust in different scenarios in terms of success rate and speed to solve motion planning queries.

The sweep line method in a single C-slice avoids traditional collision detection computation in generating collision-free samples. The vertices computed in each C-slice are automatically guaranteed to be safe. With the enhancement step, more vertices within each free segment can be generated. The added new vertices are closer to the adjacent free segment than the existing middle point, making it possible to circumvent obstacles compared to directly connecting two middle points. This step makes the vertex generation process more robust since more possible valid edges can be connected. Moreover, when connecting an edge between two vertices within one C-slice, the whole edge is checked for intersections with C-obstacle boundaries (if a straight-line connection is considered). This is a continuous way of performing validity check, since no interpolation along the edge is required. 

With the roadmap refinement process, the portion of free space represented by the collision-free intervals of each sweep line increases with higher resolutions. Each C-slice can be explored uniformly along the sweeping direction and completely within a certain resolution parameter. The initial resolution parameter set by users might not be enough to find a valid solution. But with this refinement step, the free space can be explored in an adaptive way, making the proposed algorithms more robust in dealing with resolution errors.

The ``bridge C-slice'' adds another C-slice to the whole roadmap, which doubles the total number of slices. However, it simplifies the edge validation process. In each bridge C-slice, only the center point of each enlarged ellipsoidal void for robot part is checked with C-obstacles. Computing a path in the bridge C-slice can be viewed as a projection of the $\SE(3)$ (or $\SE(3) \times (S^1)^{n}$) motion sequence of the robot onto a path for translational motion of the enlarged void in $\IR^3$. The validation process still involves interpolations between two $\SO(3)$ (or $\SO(3) \times (S^1)^{n}$) configurations. But they are only computed once before connecting two C-slices. 

The deterministic nature of HRM makes it stable over different benchmark trials on the same planning scene. The Prob-HRM planner, on the other hand, integrates the shining features of the probabilistic ideas in sampling-based algorithms. Comparing to HRM, the number of robot shapes sampled in Prob-HRM is unknown \textit{a priori}. But as shown in the benchmark results, the final numbers of C-slices are within a tractable range. This is mainly because that Prob-HRM still preserves the deterministic nature when exploring each C-slice, which increases the chance of identifying difficult regions. The collaboration with sampling-based planners avoids the dimensionality explosions for higher degrees-of-freedom robots, making our framework extendable to wider and more complicated tasks.

%%%%%%%%%%%%%%%%%%%%%%%%%%%%%%%%%%%%%%%%%%%%%%%%%%%%%%%%%%%%%%%%%%%%%%%%%%%%%%%%%%%
\subsection{Limitations}
There are also some limitations of the proposed framework. Firstly, the geometry of the robot parts is limited to ellipsoids. The Minkowski sums are exact only when one of the bodies is an ellipsoid. For other geometric representations, such as polyhedra and point cloud, a fitting process is required before running the planner. Also, the meshed surface of the exact Minkowski sum in the sweep-line process introduces another level of approximation errors.

The computation of tightly-fitted ellipsoid (TFE) in the bridge C-slice is conservative in the sense that some free space will be lost when the robot parts are enlarged. This enlargement scarifies the completeness of the planner. However, the efficiency using the bridge C-slice is significant based on the ablation study and benchmark results. Also, when the distance between two C-slices is smaller, the TFE encloses each robot part tighter, resulting in losing less free space. 

The current HRM and Prob-HRM are both effective when the robot motions are dominated by translations. But they are not advantageous for robots with a fixed base such as manipulators. Prob-HRM can possibly be used to solve problems with pure rotational motions. In this case, useful operations within a single C-slice might be very limited, since no translational connections can be made. When the robot base is fixed, Prob-HRM is equivalent to a pure sampling-based planner. In this case, the proposed closed-form Minkowski operations and the sweep line method can be used to generate valid vertices during the C-space exploration. And the ``bridge C-slice'' method can be applied as the transition validity checker between adjacent C-slices.

% Despite the efficiency of the HRM and Prob-HRM in solving narrow passage problems, t

% Conclusion
\section{Conclusion} 
\label{sec:conclude}
This article proposes a path planning framework based on the closed-form characterization of Minkowski sum and difference. The important ``narrow passage'' problem can be solved efficiently by the proposed extended Highway RoadMap (HRM) planner. Collision-free configurations are generated directly by a ``sweep line'' process. And connections between two configurations with the same rotational components can be validated without interpolations. Configurations with different rotational components are connected through a novel ``bridge C-slice'' method using the sweep volume of enlarged ellipsoidal voids. A new hybrid probabilistic variant, \ie Prob-HRM, is then proposed to solve higher dimensional problems. It combines the efficient explicit descriptions of C-space and the effectiveness of random sampling. This hybrid idea can thereby achieve better performance in higher dimensional (articulated robot) motion planning problems in cluttered environments with narrow passages.

%%%%%%%%%%%%%%%%%%%%%%%%%%%%%%%%%%%%%%%%%%%%%%%%%%%%%%%%%%
\section*{Acknowledgement}
The authors would like to thank Dr. Yan Yan, Dr. Yuanfeng Han and Mr. Chan Wee Kiat for useful discussions and helps on the physical experiments. This work was performed under National Research Foundation, Singapore, under its Medium Sized Centre Programme - Centre for Advanced Robotics Technology Innovation (CARTIN), subaward R-261-521-002-592, Singapore MOE Tier 1 grant R-265-000-655-114, National University of Singapore (NUS) Startup grants R-265-000-665-133 and R-265-000-665-731 and NUS Faculty Board funds C-265-000-071-001. When the authors were at Johns Hopkins University, they were supported by National Science Foundation of U.S. grant IIS-1619050 and U.S. Office of Naval Research award N00014-17-1-2142. The ideas expressed in this paper are solely those of the authors.

%%%%%%%%%%%%%%%%%%%%%%% Reference %%%%%%%%%%%%%%%%%%%%%%%%
\bibliographystyle{IEEEtran}
\bibliography{reference}

\end{document}